%% file: egbib.tex
\documentclass[10pt,twocolumn,letterpaper]{article}

\usepackage{iccv}
\usepackage{times}
\usepackage{epsfig}
\usepackage{graphicx}
\usepackage{amsmath}
\usepackage{amssymb}
\usepackage{blindtext}
\usepackage{caption}
\usepackage{subfig}
\usepackage{multirow}
\usepackage{booktabs}
\usepackage{verbatim}
\usepackage{color, colortbl}
\usepackage{xspace, xcolor}
\usepackage{cite}
\usepackage[export]{adjustbox}

\input{notations}

\definecolor{Gray}{gray}{0.9}

\usepackage[pagebackref=true,breaklinks=true,letterpaper=true,colorlinks,bookmarks=false]{hyperref}

 \iccvfinalcopy % *** Uncomment this line for the final submission

\def\iccvPaperID{1042} % *** Enter the ICCV Paper ID here

\ificcvfinal\fi
\begin{document}

%%%%%%%%% TITLE
%Focus For Free in Local and Global Counting
\title{Counting with Focus for Free}

\author{Zenglin Shi, Pascal Mettes, and Cees G. M. Snoek \\University of Amsterdam}

%\author{Zenglin Shi\\
%University of Amsterdam\\
%{\tt\small z.shi@uva.nl}
%\and
%University of Amsterdam\\

\maketitle
%%%%%%%%% ABSTRACT
\input{abstract}
\input{introduction}
\input{related-work}

\input{model}
\input{experiment}
\input{conclusion}

%CS: Needed for DIVA/final version
%\vspace{2mm}
%%%%
%{
%	\small
%	\textbf{Acknowledgments}
%	Supported by the Intelligence Advanced Research Projects Activity (IARPA) via Department of Interior/Interior Business Center (DOI/IBC) contract number D17PC00343. The U.S. Government is authorized to reproduce and distribute reprints for Governmental purposes notwithstanding any copyright annotation thereon. Disclaimer: The views and conclusions contained herein are those of the authors and should not be interpreted as necessarily representing endorsements, either expressed or implied, of IARPA, DOI/IBC, or the U.S. Government.
%}
{\small
\bibliographystyle{ieee}
\bibliography{egbib}
}
\clearpage
\input{appendix}

\end{document}

%% file: notations.tex
\usepackage{xspace}
\def\eg{\textit{e.g.}}
\def\ie{\textit{i.e.}}

\def\etal{\textit{et al. }}

%% file: abstract.tex
\begin{abstract}
This paper aims to count arbitrary objects in images. The leading counting approaches start from point annotations per object from which they construct density maps. Then, their training objective transforms input images to density maps through deep convolutional networks. We posit that the point annotations serve more supervision purposes than just constructing density maps. We introduce ways to repurpose the points for free. First, we propose supervised focus from segmentation, where points are converted into binary maps. The binary maps are combined with a network branch and accompanying loss function to focus on areas of interest. Second, we propose supervised focus from global density, where the ratio of point annotations to image pixels is used in another branch to regularize the overall density estimation. To assist both the density estimation and the focus from segmentation, we also introduce an improved kernel size estimator for the point annotations. Experiments on six datasets show that all our contributions reduce the counting error, regardless of the base network, resulting in state-of-the-art accuracy using only a single network. Finally, we are the first to count on WIDER FACE, allowing us to show the benefits of our approach in handling varying object scales and crowding levels. Code is available at \url{https://github.com/shizenglin/Counting-with-Focus-for-Free}
\end{abstract}

%% file: introduction.tex
%===============================
\section{Introduction}
%===============================
This paper strives to count objects in images, whether they are people in crowds~\cite{zhang2016single, yang2016wider,idrees2018composition}, cars in traffic jams~\cite{guerrero2015extremely} or cells in petri dishes~\cite{Marsden_2018_CVPR}. The leading approaches for this challenging problem count by summing the pixels in a density map \cite{lempitsky2010learning} as estimated with a convolutional neural network, \eg~\cite{Cao_2018_ECCV,Li_2018_CVPR,Issam2018Where,Marsden_2018_CVPR}. While this line of work has shown to be effective, the rich source of supervision from the point annotations is only used to construct the density maps for training. The premise of this work is that point annotations can be repurposed to further supervise counting optimization in deep networks, for free.

\begin{figure}[t]
\includegraphics[width=\linewidth]{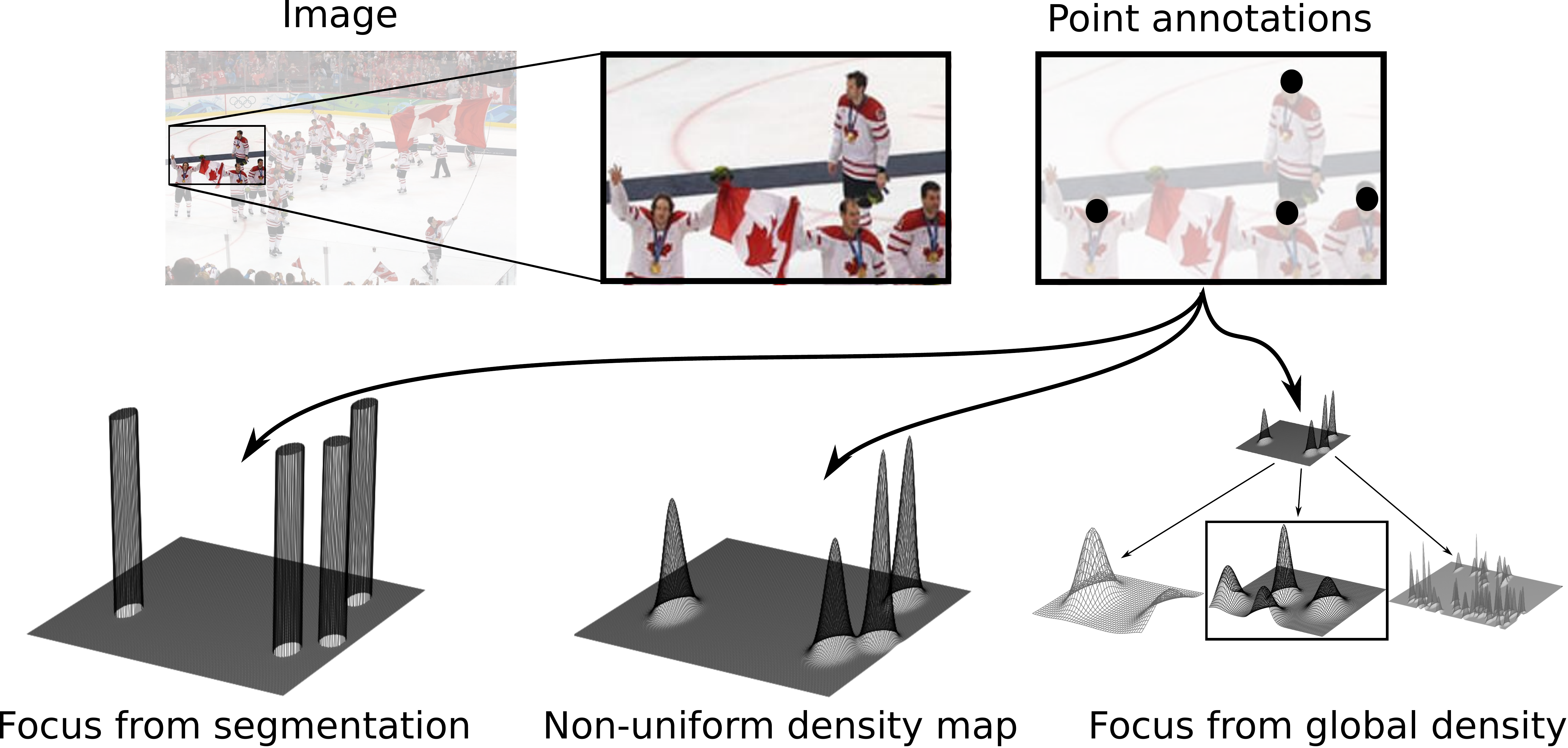}
\caption{\textbf{Focus for free} in counting. From point supervision, we learn to obtain a focus from segmentation, a focus from global density, and an improved density maps. Combined, they result in better counting estimation irrespective of the base network.}
\label{fig:fig1}
\vspace{-4mm}
\end{figure}

The main contribution of this paper is summarized in Figure~\ref{fig:fig1}. Besides creating density maps, we show that points can be exploited as free supervision signal in two other ways. The first is focus from segmentation. From point annotations, we construct binary segmentation maps and use them in a separate network branch with an accompanying segmentation loss to focus on areas of interest only. The second is focus from global density. The relative amount of point annotations in images is used to focus on the global image density through another branch and loss function. Both forms of focus are integrated with the density estimation in a single network trained end-to-end with a multi-level loss.
In standard attention mechanisms~\cite{hu2018squeeze,woo2018cbam, Liu_2018_CVPR_decidenet,KangBMVC2018}, the weighing map is indirectly learned from a task-specific objective, \eg image classification or object counting. We also rely on task-specific supervision, but we explicitly add novel supervised network branches for the segment and density weighting maps. We derive the necessary supervision from provided point annotations and name it focus for free.

%
%Different from standard attention~\cite{xu2015show, chen2017sca,Liu_2018_CVPR_decidenet,KangBMVC2018}, where a form of focus needs to be learned for the task at hand and the learned weighing map implicitly guides the network to focus on task-relevant features, our proposed focus learns weighing maps with a specific supervision derived for free from point annotations. Focus for free allows the counting network to explicitly emphasize meaningful features and suppress undesired ones.

Overall, we make three contributions in this paper: (\textit{i}) We propose supervised focus from segmentation, a network branch which guides the counting network to focus on areas of interest. The supervision is obtained from the already provided point annotations. (\textit{ii}) We propose supervised focus from global density, a branch which regularizes the counting network to learn a matching global density. Again the supervision is obtained for free from the point annotations. (\textit{iii}) We introduce a new kernel density estimator for point annotations with non-uniform point distributions. For the deep network, we design an improved encoder-decoder network to deal with varying object scales in images. Experimental evaluation on six counting datasets shows the benefits of our focus for free, kernel estimation, and end-to-end network architecture, resulting in state-of-the-art counting accuracy. To further demonstrate the potential of our approach for counting under varying object scales and crowding levels, we provide the first counting results on WIDER FACE, normally used for large-scale face detection~\cite{yang2016wider}.

%% file: related-work.tex
%===============================
\section{Related Work}
%===============================
\noindent
\textbf{Density-based counting.}
%\cs{I don't understand this sentence:}
%
%Compared to only the total count as a single value, density maps preserve spatial information, making them more suitable for counting \cite{lempitsky2010learning,zhang2016single,Shi_2018_CVPR}.
%
Deep convolutional networks are widely adopted for counting by estimating density maps from images. Early works, \eg~ \cite{zhang2015cross,zhang2016single,onoro2016towards,sindagi2017generating}, advocate a multi-column convolutional neural network to encourage different columns to respond to objects at different scales. Despite their success, these types of networks are hard to train due to structure redundancy \cite{Li_2018_CVPR} and conflicts resulting from optimization among different columns \cite{Shen_2018_CVPR, Sam_2018_CVPR}.

Due to their architectural simplicity and training efficiency, single column deep networks have received increasing interest \eg~\cite{Li_2018_CVPR,Cao_2018_ECCV,Liu_2018_CVPR_Leveraging,Shi_2018_CVPR, liu2018crowd}. Cao \etal \cite{Cao_2018_ECCV} , for example, propose an encoder-decoder network to predict high-resolution and high-quality density maps using a scale aggregation module. Li \etal \cite{Li_2018_CVPR} combine a VGG network with dilated convolution layers to aggregate multi-scale contextual information. Liu \etal \cite{Liu_2018_CVPR_Leveraging} rely on a single network by leveraging abundantly available unlabeled crowd imagery in a learning-to-rank framework. Shi \etal \cite{Shi_2018_CVPR} train a single VGG network with a deep negative correlation learning strategy to reduce the risk of over-fitting. We also employ single column networks, but rather than focusing solely on density map estimation, we repurpose the point annotations in multiple ways to improve counting.

Recently, multi-task networks have shown to reduce the counting error \cite{sam2017switching,ranjan2018iterative,Shi2018vlad,Shen_2018_CVPR,Sam_2018_CVPR,Liu_2018_CVPR_decidenet,idrees2018composition}. Sam \etal \cite{sam2017switching}, for example, train a classifier to select the optimal regressor from multiple independent regressors for particular input patches. Ranjan \etal \cite{ranjan2018iterative} rely on one network to predict a high resolution density map and a helper-network to predict a density map at a low resolution. 
%Multi-task networks obtain impressive counting accuracy at the expense of complicated learning, since it requires individual tuning for each sub-network. 
In this paper, we also investigate counting from a multi-task perspective, but from a different point of view. We posit that the point annotations serve more purposes than just constructing density maps, and we propose network branches with supervised focus from segmentation and global density to repurpose the point annotations for free. Our focus for free benefits counting regardless of the base network, and is complementary to other state-of-the-art solutions. 

%\cs{Need to cite: S. Zhang, G.Wu, J. P. Costeira, and J. M. Moura. Understanding traffic density from large-scale web camera data. CVPR, 2017. Also update table with SotA on TRANCOS}
 
\textbf{Counting with attention.}
Attention mechanisms \cite{xu2015show} have enabled progress in a wide variety of computer vision challenges \cite{chen2017sca,girdhar2017attentional,LiCVIU18,Zhang_2018_CVPR,Zhu_2018_CVPR}. Soft attention is the most widely used since it is differentiable and thus can be directly incorporated in an end-to-end trainable network. The common way to incorporate soft attention is to add a network branch with one or more hidden layers to learn an attention map which assigns different weights to different regions of an image. Spatial and channel attention are two well explored types of soft attention~\cite{chen2017sca,woo2018cbam}. Spatial attention learns a weighting map over the spatial coordinates of the feature map, while channel attention does so for the feature channels of the map.

A few works have investigated density-based counting with spatial attention~\cite{Liu_2018_CVPR_decidenet,KangBMVC2018,hossain2019crowd}. Liu \etal \cite{Liu_2018_CVPR_decidenet}, for example, estimate the density of a crowd by generating separate detection- and regression-based density maps. They fuse these two density maps guided by an attention map, which is implicitly learned together with the density map regression loss. While we share the notion of assisting the density-based counting with a focus, we show in this work that such an attention does not need to be learned from scratch and instead can be derived from the existing point annotations.
%In this way, the learned weighting map can not explicitly guide the network to focus on task relevant regions. To combat this limitation, we propose to provide supervision directly on the attention of the network.
More specifically, we construct a segmentation map and a global density derived from the ground-truth annotated points as two additional, yet free, supervision signals for better counting.

%% file: model.tex
%===============================
\section{Focus for Free} \label{model}
%===============================

\begin{figure*}[!t]
\centering
\includegraphics[width=0.99\linewidth]{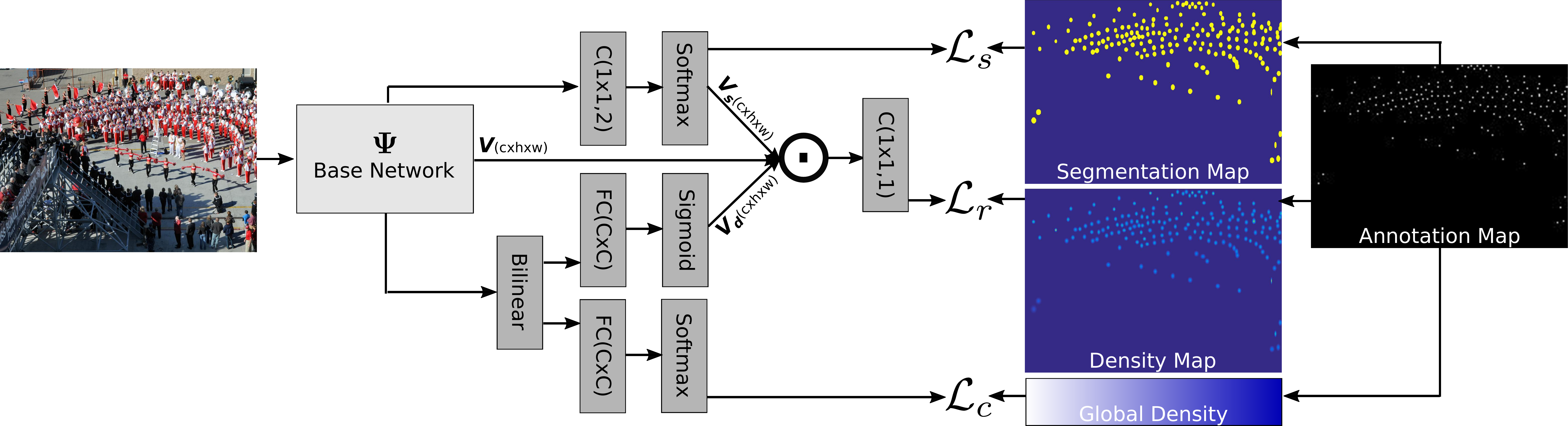}
\caption{\textbf{Overview of our approach}. Top branch: focus from segmentation learns a focus map $V_s$ with the aid of a segmentation map (Section \ref{sub_segment}). Bottom branch: focus from global density learns a focus map $V_d$ with the aid of a global density (Section \ref{sub_global}). Both supervision signals are obtained from the same point-annotations, for which we introduce an improved kernel estimator (Section \ref{sub_kernel}). Both branches with focus for free are integrated with the output of a base network by element-wise multiplication and end-to-end optimized through a multi-level loss (Section \ref{sub_optimization}).} %Irrespective of base network (Section \ref{sub_cross_network})}. 
%\psmm{final sentence does not sound like a requirement and can go in my opinion.}} % Network details in respective section.
%
%C represents the convolution layer with parameters $(k \times k, c)$, where $k \times k$ is the kernel size, $c$ is the number of channels, FC denotes fully-connected layer, and Bilinear denotes bilinear pooling. \cs{TO DISCUSS}} %$\mathcal{L}_s$ and $\mathcal{L}_c$ denote the loss functions for object segmentation and global density level prediction, respectively. The feature maps $V$ extracted from the base network $\psi$ are forwarded to learn a focus map $V_s$ with the aid of a segmentation map, and a focus map $V_d$ with the aid of a global density. Both supervision signals are obtained for free from the provided point-annotations. Then the two types of focus is integrated with the output of base network by element-wise multiplication, and the resulting output is fed into a $1\times1$ convolution layer to generate a density map. Better viewed in color.}
\label{fig_focus}
\vspace{-4mm}
\end{figure*}

We formulate the counting task as a density map estimation problem, see \eg \cite{lempitsky2010learning,zhang2016single,Shi_2018_CVPR}. Given $N$ training images $\{(X_i,\mathcal{P}_i)\}_{i=1}^{N}$, with $X_i \subset \mathcal{X}$ the input image and $\mathcal{P}_i$ a set of point annotations, one for each object, we use the point annotations to create a ground-truth density map by convolving the points with a Gaussian kernel,
\begin{equation}
D_i(p)=\sum_{P\in \mathcal{P}_i}\mathcal{N}(p|\mu=P,\sigma_P^2),
\end{equation}
where $p$ denotes a pixel location, $P$ denotes a single point annotation and $\mathcal{N}(p|\mu=P,\sigma_P^2)$ is a normalized Gaussian kernel with mean $P$ and an isotropic covariance $\sigma_P^2$. The global object count $T_i$ of image $X_i$ can be obtained by summing all pixel values within the density map $D_i$, \ie, $T_i=\sum_{p \in X_i}D_i(p)$. Learning a transformation from input images to density maps is done through deep convolutional networks. Let $\Psi(X): \mathbb{R}^{3 \times W \times H} \mapsto \mathbb{R}^{W \times H}$ denote such a mapping given an arbitrary deep network $\Psi$ for image $X$, with $W$ and $H$ the width and height of the image. In this paper, we investigate two ways that repurpose the point annotations to help supervising the network $\Psi$ from input images to density maps. An overview of our approach, in which multiple branches are combined on top of a base network, is shown in Figure \ref{fig_focus}.

%===============================
\subsection{Focus from segmentation}
\label{sub_segment}
%===============================
The first way to repurpose the point annotations is to provide a spatial focus. Intuitively, pixels that are within a specific range of any point annotation should be of high focus, while pixels in undesired regions should be mostly disregarded. In the standard setup where the optimization is solely dependent on the density map, each pixel counts equally to the network loss. Given that only a fraction of the pixels are near point annotations, the loss will be dominated by the majority of irrelevant pixels. To overcome this limitation, we reuse the point annotations to create a binary segmentation map and exploit this map to provide the focused supervision through a stand-alone loss function.

\textbf{Segmentation map.} The binary segmentation map is obtained as a function of the point annotations and their estimated variance. The binary value for each pixel $p$ in training image $i$ is determined as:
\begin{equation}
S_i(p) =
\begin{cases}
1 & \text{if  $\exists_{P \in \mathcal{P}_i} \big( ||p - P||^2 \leq \sigma_P^2$} \big),\\
0 & \text{otherwise.}
\end{cases}
\label{eq:segfocus}
\end{equation}
Equation~\ref{eq:segfocus} states that a pixel $p$ obtains a value of one if at least one point $P$ is within its variance range $\sigma_P$ as specified by a kernel estimator.

\textbf{Segmentation focus.} Let $V \in \mathbb{R}^{C \times W \times H}$ denote the output of the base network. We add a new branch on top of the network denoted as $\mathcal{F}_s$ with network parameters $\theta_s$. Furthermore, let $\theta_n$ denote the parameters of the base network. We propose a per-pixel weighted focal loss~\cite{lin2017focal} to obtain a supervised focus from segmentation for input image $X$:
\begin{equation}
\begin{split}
\mathcal{L}_s(X;\theta_n,\theta_s) = 
 \sum_{{l} \in \{0,1\}} -\alpha{^l} S^l \\ (1-\mathcal{F}_{s}(X; \theta_n,\theta_s))^{\gamma_s} log(\mathcal{F}_{s}(X; \theta_n,\theta_s)),
\label{eq:focus-seg}
\end{split}
\end{equation}
where $\alpha{^l}=1-\frac{|S^l|}{|S|}$. The focal parameter $\gamma_s$ is set to 2 throughout this network, as recommended by~\cite{lin2017focal}.
The segmentation branch is visualized at the top of Figure \ref{fig_focus}. 

\textbf{Network details.} After the output of the base network, we perform a $1 \times 1$ convolution layer with parameters $\theta_{s} \in \mathbb{R}^{C \times 2 \times 1 \times 1}$, followed by a softmax function $\delta$ to generate a per-pixel probability map $\mathbb{P}_i = \delta(\theta_{s} V) \in \mathbb{R}^{2 \times W \times H}$. From this probability map, the second value along the first dimension represents the probability of each pixel being part of the segmentation foreground. We furthermore tile this slice $C$ times to construct a separate output tensor $V_s \in \mathbb{R}^{C \times W \times H}$, which will be used in the density estimation branch itself.

%===============================
\subsection{Focus from global density}
\label{sub_global}
%===============================
Next to a spatial focus, point annotations can also be repurposed by examining their context. It is well known that low density crowds exhibit coarse texture patterns while high density crowds exhibit very fine texture patterns. Here, we exploit this knowledge for the task of counting. Given a network output $V \in \mathbb{R}^{W \times H \times C}$, we employ a bilinear pooling layer \cite{lin2015bilinear,gao2016compact} to capture the feature statistics in a global context, which is known to be particularly suitable for texture and fine-grained recognition \cite{lin2015bilinear,gao2016compact}. In this work, we match global contextual patterns to the distribution of points in training images to obtain a supervised focus from global density.

\textbf{Global density.}
For patch $j$ in training image $i$, its global density is given as:
\begin{equation}
G_{j,i} = \frac{|\mathcal{P}_{j,i}|}{L},
\end{equation}
where $|\mathcal{P}_{j,i}|$ denotes the number of point annotations in patch $j$ and $L$ denotes the global density step size, which is computed for a dataset as:
\begin{equation}
L = \left \lfloor{ \max_{i=1,..,N} \Big (\frac{|\mathcal{P}_{i}|}{Z_i} \cdot Z_{j,i}\Big) / M}\right \rfloor + 1,
\end{equation}
with $Z_i$ and $Z_{j,i}$ the number of pixels in image $i$ and patch $j$ respectively. Intuitively, the step size computes the maximum global density over image patches and $M$ states how many global density levels are used overall.

\textbf{Global density focus.} With $V \in \mathbb{R}^{C \times W \times H}$ again the output of the base network, we add a second new branch $\mathcal{F}_c$ with network parameters $\theta_c$. We propose the following global density loss function:
\begin{equation}
 \begin{split}
\mathcal{L}_c(X_;\theta_n,\theta_c) = 
\sum_{{l} \in \{0,1,..,M\}} -G^l \\ (1-\mathcal{F}_{c}(X; \theta_n,\theta_c))^{\gamma_c} log(\mathcal{F}_{c}(X; \theta_n,\theta_c)),
\label{eq:focus-compactness}
\end{split}
\end{equation}
where $\gamma_c$ is set to 2 as well. The above loss function aims to match the global density of the estimated density map with the global density of the ground truth density map. The corresponding global density branch is visualized at the bottom of Figure \ref{fig_focus}. 

\textbf{Network details.} For network output $V$, we first perform an outer product $B = VV^T \in \mathbb{R}^{C \times C}$, followed by a mean pooling along the second dimension to aggregate the bilinear features over the image, \ie~$\hat{B} = \frac{1}{C} \sum_{i=1}^{C} B[:,i] \in \mathbb{R}^{C \times 1}$. The bilinear vector $\hat{B}$ is $\ell_2$-normalized, followed by signed square root normalization, which has shown to be effective in bilinear pooling~\cite{lin2015bilinear}. Then we use a fully connected layer with parameters $\theta_c \in \mathbb{R}^{C \times M}$ followed by a softmax function $\delta_c$ to make individual prediction $\mathbb{C} = \delta_c(\theta_c \hat{B}) \in \mathbb{R}^{M \times 1}$ for the global density. Furthermore, another fully-connected layer with parameters $\theta_d \in \mathbb{R}^{C \times C}$ followed by sigmoid function $\delta_d$ also on top of the bilinear pooling layer is added to generate global density focus output $\mathbb{D} = \delta_d(\theta_d \hat{B}) \in \mathbb{R}^{C \times 1}$. We note that this results in a focus over the channel dimensions, complementary to the focus over the spatial dimensions from segmentation. Akin to the focus from segmentation, we tile the output vector into $V_d \in \mathbb{R}^{C \times W \times H}$, also to be used in the density estimation branch.

%===============================
\subsection{Non-uniform kernel estimation}
\label{sub_kernel}
%===============================
% To discuss, how can we claim our kernel non-uniform?
Both the density estimation itself and the focus from segmentation require a variance estimation for each point annotation, where the variance corresponds to the size of the object. Determining the variance $\sigma_P$ for each point $P$ is difficult because of object-size variations caused by perspective distortions. A common solution is to estimate the size (\ie~the variance) of an object as a function of the $K$ nearest neighbour annotations, \eg~the Geometry-Adaptive Kernel of Zhang \etal \cite{zhang2016single}. However, this kernel is effective only under the assumption that objects in images are uniformly distributed, which typically does not happen in counting practice. As such, we introduce a simple kernel that estimates the variance of a point annotation $P$ by splitting an image into local regions:
\begin{equation}
\sigma_P = \frac{1}{|R_{(w,h)}|} \sum_{a \in R_{(w,h)}}\beta \bar{d_a}, \quad  \bar{d_a}=\frac{1}{K} \sum_{k=1}^{K}d_{k,a}
\end{equation}
where $w$ and $h$ are the hyper-parameters which determine the range of point annotation $P$-centered local region $R$, and we set their value to one-eighth of image size in our experiments. $a$ denotes an arbitrary point annotation located in $R$. $|R_{(w,h)}|$ means the number of $p$. $\bar{d_p}$ indicates the average distance between annotated point $p$ and its $k$ nearest neighbors, and $\beta$ is a user-defined hyper-parameter.
By estimating the variance of point annotations locally, we no longer have to assume that points are uniformly distributed over the whole image.

%===============================
\subsection{Architecture and optimization}
\label{sub_optimization}
%===============================
\textbf{Network.}
To maximize the ability to focus and use the most accurate kernel estimation, we want the network output to be of the same width and height as the input image. Recently, encoder-decoder networks have been transferred from other visual recognition tasks~\cite{yu2017dilated,lin2017feature} to counting ~\cite{Shen_2018_CVPR,Zhang_2017_CVPR,ranjan2018iterative,Cao_2018_ECCV}. We found that to make the encoder-decoder architectures better suited for counting, the wide variation in object-scale under perspective distortions needs to be addressed. As such, in our encoder-decoder architecture a distiller module is added between the step from encoder to decoder. The purpose of this module is to aggregate multi-level information from the encoder by distilling the most vital information for counting.

For the encoder, we make the original dilated residual network \cite{yu2017dilated} suitable for our task by changing the channel of the feature maps after level 4 from 256/512 to 96 to reduce the model's parameters for the sake of avoiding over-fitting, given the low amount of training examples in counting. After the encoder, the distiller module fuses the features from level 4, 5, 7 and 8 in the encoder module by using skip connections and a concatenation operation. Then four convolution layers are used to further process the fused features to obtain a more compact representation. The reason why we do not fuse the features from level 6 is that level 6 comprises convolution layers with large dilation rates, which is prone to cause gridding artifacts \cite{yu2017dilated,wang2017understanding}. Compared to other works which fuse multiple networks with different kernels to deal with object-scale variations \cite{onoro2016towards, zhang2016single, sindagi2017generating}, the proposed network aggregates the features from different layers which have different receptive fields, and is much more efficient and easy to train. The decoder module uses 3 deconvolution layers with a kernel size of $4 \times 4$ and a stride size of $2 \times 2$ to progressively recover the spatial resolution. To avoid the checkerboard artifact problem caused by regular deconvolutional operation \cite{odena2016deconvolution,wang2017understanding}, we add two convolution layers after each deconvolution layer.
%Although adding skip connections between encoder and decoder modules has shown to be helpful to recover spatial information and preserve the fine details in many pixel-wise prediction tasks \cite{RFB15a, long2015fully, mao2016image}, we prefer to avoid these connections for our counting task because too much background noise from the low level features in the encoder will propagate to the decoder.
%
We provide a detailed ablation on the 
encoder-distiller-decoder network in the supplementary material.

\textbf{Multi-level loss.}
The final counting network with a focus for free contains three branches, $\mathcal{F}_r$ for the pixel-wise density estimation, $\mathcal{F}_s$ for the binary segmentation, and $\mathcal{F}_c$ for the global density prediction. Let $(\theta_n,\theta_r,\theta_s,\theta_c,\theta_d)$ denote the network parameters for the base network and the branches. For the density estimation, we first combine the outputs of the base network $V$ with the tiled outputs $V_s$ and $V_d$ from the focus for free. We fuse the three sources of information by element-wise multiplication and feed the fusion to a $1 \times 1$ convolution layer with parameters $\theta_{r} \in \mathbb{R}^{C \times 1 \times 1 \times 1}$, resulting in an output density map.

For the density estimation, the $L2$ loss is a common choice, but it is also known to be sensitive to outliers, which hampers generalization \cite{belagiannis2015robust}. We prefer to learn the density estimation branch by jointly optimizing the $L2$ and $L1$ loss, which adds robustness to outliers:
\begin{equation}
 \begin{split}
\mathcal{L}_r(X;\theta_n,\theta_r,\theta_d) = \frac{1}{2} \parallel \mathcal{F}_r(X; \theta_n,\theta_r,\theta_d)-Y \parallel_{2}^{2}+  \\
\parallel \mathcal{F}_r(X; \theta_n,\theta_r,\theta_d)-Y \parallel_{1},
\end{split}
\label{eq:loss-mainbranch}
\end{equation}
where $Y$ denotes the ground truth density map. Empirically, we also find that this combined loss is preferred over only using the $L1$ or $L2$ loss. The loss functions of the three branches are summed to obtain the final objective function:
\begin{equation}
 \begin{split}
\mathcal{L}(X;\theta_n,\theta_r, \theta_s, \theta_c,\theta_d) =
\lambda_r \mathcal{L}_r(X;\theta_n,\theta_r,\theta_d)+ \\ \lambda_s \mathcal{L}_s(X;\theta_n,\theta_s)+ \lambda_c \mathcal{L}_c(X;\theta_n,\theta_c),
\label{eq:final-loss}
\end{split}
\end{equation}
where $(\lambda_r,\lambda_s,\lambda_c)$ denote the weighting parameters of the different loss functions. Throughout this work these parameters are set to $(1,10,1)$, since the loss values of the segmentation branch are typically an order of magnitude lower than the others.

%% file: experiment.tex
%===============================
\section{Experimental Setup}
%===============================

%-------------------------------
\subsection{Datasets} 
%-------------------------------

\textbf{ShanghaiTech} \cite{zhang2016single} consists of 1198 images with 330,165 people. This dataset is divided into two parts: \textbf{Part\_A} with 482 images in which crowds are mostly dense (33 to 3139 people), and \textbf{Part\_B} with 716 images, where crowds are sparser (9 to 578 people). Each part is divided into a training and testing subset as specified in \cite{zhang2016single}. 
\textbf{TRANCOS} \cite{guerrero2015extremely} contains 1,244 images from different roads to count vehicles, varying from 9 to 105. We train on the given training data (403 images) and validation data (420 images) without any other datasets, and we evaluate on the test data (421 images). 
\textbf{Dublin Cell Counting (DCC)} \cite{Marsden_2018_CVPR} is a cell microscopy dataset, consisting of 177 images, with a cell count from 0 to 100. For training 100 images are used, the remaining 77 form the test set.
\textbf{UCF-QNRF} \cite{idrees2018composition} is a recent large-scale crowd dataset, consisting of 1,535 images, with the count ranging from 49 to 12,865. For training 1201 images are used, the remaining 334 form the test set.
\textbf{WIDER FACE}~\cite{yang2016wider} is a face detection benchmark. In this paper, we repurpose it for counting as a complementary crowd dataset. Compared to ShanghaiTech~\cite{zhang2016single} and UCF-QNRF~\cite{idrees2018composition}, WIDER FACE is more challenging due to large variations in scale, occlusion, pose, and background clutter. Moreover, it contains more images, in total $32,203$, divided in 40\% training, 10\% validation and 50\% testing. The ground truth of the test set is unavailable, so we report on the validation set. Each face is annotated by a bounding box, instead of a point, which enables us to evaluate our kernel estimator and allows for ablation under varying object scales and crowding levels.

%-------------------------------
\subsection{Implementation details} 
%-------------------------------
\textbf{Pre-processing.} For all datasets, we normalize the input RGB images by dividing all values by 255. During training, we augment the images by randomly cropping $128 \times 128$ patches. No cropping is performed during testing.

\textbf{Network.} We implement our method with TensorFlow on a machine with a single GTX 1080 Ti GPU. The network is trained using Adam with a mini-batch of 16. We set the $\beta_1$ to 0.9, $\beta_2$ to 0.999 and the initial learning rate to 0.0001. Training is terminated after a maximum of 1000 epochs.

\textbf{Kernel computation.} For datasets with dense objects, \ie~ShanghaiTech Part\_A, TRANCOS and UCF-QNRF, we use our proposed kernel with $\beta=0.3$ and $k=5$. For ShanghaiTech Part\_B and DCC, we set the Gaussian kernel variance to $\sigma=5$ and $\sigma=10$ respectively, following~\cite{Shi_2018_CVPR,Li_2018_CVPR}. For WIDER FACE, we obtain the Gaussian kernel variance by leveraging the box annotations.
%\psmm{Do you not use our kernel here? I assumed that you only use the ground truth variance for ablation studies.} 
For the focus from global density, we use $M=8$ density levels for ShanghaiTech Part\_A and UCF-QNRF, and 4 for the other datasets. 

%-------------------------------
\subsection{Evaluation metrics} 
%-------------------------------

\textbf{Count error.} We report the Mean Absolute Error (MAE) and Root Mean Square Error (RMSE) metrics given count estimates and ground truth counts~\cite{zhang2015cross,zhang2016single,Shi_2018_CVPR}.
%
%\textbf{Local count accuracy.} 
Since these global metrics ignore where objects have been counted, we also report results using the Grid Average Mean absolute Error (GAME) metric.~\cite{guerrero2015extremely}. GAME aggregates count estimates over local regions as:
%
%\begin{equation}
$GAME(L) = \frac{1}{N}\cdot \sum_{n=1}^{N}( \sum_{l=1}^{4^L}|(y_{n}^l- \tilde{y}_{n}^l)|),$
%\end{equation}
%
with $N$ the number of images and $y_{n}^l$ and $\tilde{y}_{n}^l$ the ground truth and the estimated counts in a region $l$ of the $n^{th}$ image. $4^L$ denotes the number of grids, non-overlapping regions which cover the full image. When $L$ is set to $0$ the GAME is equivalent to the MAE. 

\textbf{Density map quality.} Finally, we report PSNR (Peak Signal-to-Noise Ratio) and SSIM (Structural Similarity in Image \cite{wang2004image}), to evaluate the quality of the predicted density maps. We only report these on ShanghaiTech Part\_A because they are not commonly reported on the other datasets.
%\cs{Need to explain we report this only for ShanghaiTech part A?}

%===============================
\section{Results}
%===============================
%\psmm{To discuss: which experiment will have which qualitative analysis?}
%MAE, ShanghaiTech Part\_A, WIDER FACE, Base network, spatial-attention, channel-attention
%===============================
\subsection{Focus from segmentation}
%===============================
We first analyze the effect of focus from segmentation on both ShanghaiTech Part\_A and WIDER FACE. We compare to two baselines. The first performs counting using the base network, where the loss is only optimized with respect to the density map estimation. Unless stated otherwise, the  encoder-distiller-decoder network is used as base network in all experiments. The second baseline adds a spatial attention on top of this base network, as proposed in \cite{chen2017sca}. The results are shown in Table \ref{tab_segment}. 

For ShanghaiTech Part\_A, the base network obtains an MAE of 74.8. The addition of spatial-attention increases the count error to 84.5 MAE, as it fails to emphasize relevant features. In contrast, focus from segmentation can explicitly guide the network to focus on task-relevant regions and it reduces the count error from 74.8 to 72.3 MAE.

For WIDER FACE, the box annotations allow us to perform an ablation on the accuracy as a function of the object scale.
%Due to more accurate focus on object region, our method can perform better in counting objects with varied size. To demonstrate this point, we further evaluate our method on WIDER FACE.
We define the scale levels of each image as $I_{scale}=\frac{F_s}{F_n}$, where $F_s$ and $F_n$ denote face size and face number. We sort the test images in ascending order according to their scale level. Finally, the test images are divided uniformly into three sets: small, medium and large.
%
%As shown in Table~\ref{tab_segment}, our method achieves better performance on all levels of scale.
In Table \ref{tab_segment}, we provide the results across multiple object scales. We observe that across all object scales, our approach is preferred, reducing the MAE from 4.7 (base network) and 4.8 (with spatial attention) to 4.3. The ablation also reveals why spatial attention is not very effective overall; while improvements are obtained when objects are small, spatial attention performs worse when objects are large. Segmentation focus from reused point annotations avoids such issues.

\begin{table}[t]
\small
\caption{\textbf{Effect of focus from segmentation} in terms of MAE on ShanghaiTech Part\_A and WIDER FACE. Across both datasets and across multiple object scales (small, medium, large), our approach outperforms the base network, even when adding spatial attention.}
\centering
\resizebox{\columnwidth}{!}{
\begin{tabular}{@{}lcccccccc@{}}
\toprule
 & \textbf{Part\_A}  & \multicolumn{4}{c}{\textbf{WIDER FACE}}   \\
\cmidrule(lr){2-2} \cmidrule(lr){3-6}
& overall & small & medium & large & overall \\
\hline
Base network & 74.8  &  9.2 & 2.7  & 2.2&4.7    \\
%\hline
\textbf{w/} Spatial attention \cite{chen2017sca} & 84.5  &  8.7 & 2.6  & 3.1&4.8    \\
\rowcolor{Gray}
\textbf{w/} Segmentation focus & \textbf{72.3}  &  \textbf{8.6} & \textbf{2.3}  & \textbf{2.0} &\textbf{4.3}    \\
\bottomrule
\end{tabular}}
\label{tab_segment}
\vspace{-4mm}
\end{table}

%===============================
\subsection{Focus from global density}
%===============================
Next, we demonstrate the effect of our proposed focus from global density. For this experiment, we again compare to two baselines. Apart from the base network, we compare to the channel attention of \cite{chen2017sca} and the squeeze-and-excitation block of \cite{hu2018squeeze}. For fair comparison, we replace the mean pooling used in the channel attention of \cite{chen2017sca} with bilinear pooling as used in our method for the sake of better encoding global context cues. The counting results are shown in Table \ref{tab_density}. Channel-attentions can reduce the error (from 74.8 to 73.4 and 72.6 MAE) in ShanghaiTech Part\_A compared to using the base network only, since the attention maps are learned on top of a pooling layer which encodes global context cues. Our focus from global density reduces the count error further to 71.7 MAE due to more specific focus from free supervision. 

To demonstrate that our focus has a lower error on different crowding levels, we perform a further ablation on WIDER FACE. We define the crowding levels of each image as $I_{crowding} = \frac{F_s}{I_s}*\frac{F_n}{I_s}$, where $F_s$, $I_s$, and $F_n$ denote face size, image size, and face number respectively. Then we sort the test images in ascending order according to their global density level. Finally, the test images are divided uniformly into three sets, sparse, medium and dense. As shown in Table~\ref{tab_density}, our method achieves the lowest error especially when scenes are sparse. 
%This result highlights the potential complementary nature of the two forms of focus.

\begin{table}[t]
\small
\caption{\textbf{Effect of focus from global density} in terms of MAE on ShanghaiTech Part\_A and WIDER FACE. Our approach is preferred for both datasets. The ablation study on WIDER FACE shows our focus from global density is most effective when scenes are sparse in number of objects.}
\centering
\resizebox{\columnwidth}{!}{
\begin{tabular}{@{}lcccccccc@{}}
\toprule
 & \textbf{Part\_A}  & \multicolumn{4}{c}{\textbf{WIDER FACE}}   \\
\cmidrule(lr){2-2} \cmidrule(lr){3-6}
& overall & sparse & medium & dense & overall \\
\hline
Base network & 74.8  &  2.1 & 2.5  & 9.5&4.7    \\
%\hline
\textbf{w/} Channel attention \cite{chen2017sca} & 73.4  &  1.6 & 2.3  & \textbf{7.8}&3.9    \\
\textbf{w/}  Squeeze-and-excitation \cite{hu2018squeeze}  & 72.6  &  1.7 & \textbf{1.6}  & \textbf{7.8}&3.7    \\
\rowcolor{Gray}
\textbf{w/} Global-density focus & \textbf{71.7}  &  \textbf{0.9} & \textbf{1.6}  & 8.0&\textbf{3.5}    \\
\bottomrule
\end{tabular}}
\label{tab_density}
\vspace{-4mm}
\end{table}

\begin{table*}[!t]
\small
\caption{\textbf{Effect of combined focus} in terms of MAE on ShanghaiTech Part\_A and WIDER FACE. Across dataset, object scale, and crowding level our approach outperforms the base network and a combined spatial and channel attention variant. 
%\psmm{Not entirely clear which attention and focus is here. Should the individual attentions/focuses also be in this table?}
}
\centering
%\resizebox{\columnwidth}{!}{
\begin{tabular}{@{}lcccccccccccccccc@{}}
\toprule
 & \textbf{Part\_A}  & \multicolumn{7}{c}{\textbf{WIDER FACE}}   \\
\cmidrule(lr){2-2} \cmidrule(lr){3-9}
& overall & small & medium & large & sparse & medium & dense & overall \\
\hline
Base network & 74.8  &  9.2 & 2.7  & 2.2&2.1&2.5&9.5&4.7    \\
%\hline
\textbf{w/} Spatial- \& channel-attention \cite{chen2017sca} & 71.6  &  8.3 & 2.0  & 2.3 &1.8 & 2.6 & 8.2 & 4.2   \\
\textbf{w/} Convolutional block attention module ~\cite{woo2018cbam} & 73.5  &  8.4 & 2.0  & 1.1 &1.2 & 1.8 & 8.5 & 3.8   \\
\rowcolor{Gray}
\textbf{w/} Our combined focus & \textbf{67.9}  &  \textbf{7.7} & \textbf{1.3}  & \textbf{0.6} & \textbf{0.9} & \textbf{1.4} & \textbf{7.3} & \textbf{3.2}    \\
\bottomrule
\end{tabular}%}
\label{tab_combined}
\vspace{-4mm}
\end{table*}

%===============================
\subsection{Combined focus for free}
\label{sub_cross_network}
%===============================
In the aforementioned experiments, we have shown that each focus matters for counting. In this experiment, we combine them.
%
%se two focuses for more accurate counting, in view that these two focuses aid density map estimation respectively from a local and global perspective, complementing each other. 
%
The results are shown in Table \ref{tab_combined}. The combination achieves a reduced MAE of 67.9 on ShanghaiTech Part\_A, and obtains a reduced MAE of 3.2 on WIDER FACE. We compare to alternative combined attention baselines, \ie, spatial-channel attention \cite{chen2017sca} and the convolutional block attention module \cite{woo2018cbam}. While the combinations of attentions achieves better results than using the base network alone, our approach is preferred across datasets, object scales, and crowding levels.

The focus for free is agnostic to the base network. To demonstrate this capability, we have applied it to four different base networks. Apart from our base network, we consider the multi-column network of Zhang \etal \cite{zhang2016single}, the deep single column network of Li \etal \cite{Li_2018_CVPR} and the encoder-decoder network of Cao \etal \cite{Cao_2018_ECCV}. We have re-implemented these networks and use the same experimental settings as in our base network. The results in Table \ref{tab_independent} show that our focus for free lowers the count error for all these networks on ShanghaiTech Part\_A and WIDER FACE.

\begin{table}[t]
\centering
\caption{\textbf{Focus for free across base networks} on ShanghaiTech Part\_A and WIDER FACE. Base network results based on our reimplementations. Regardless of the base network, our combined focus from segmentation and global density lowers the count error. }
\resizebox{\columnwidth}{!}{
\begin{tabular}{lcccc}
\toprule
 & \multicolumn{2}{c}{\textbf{Part\_A}} & \multicolumn{2}{c}{\textbf{WIDER FACE}}\\
 \cmidrule(lr){2-3} \cmidrule(lr){4-5}
 Network from & base & \textbf{w/} our focus & base & \textbf{w/} our focus\\
\midrule
Zhang \etal \cite{zhang2016single} & 114.5 & \textbf{110.1} & 7.1 & \textbf{6.1}\\
Cao \etal \cite{Cao_2018_ECCV} & 75.2 & \textbf{72.7} & 8.5 & \textbf{8.2}\\
Li \etal \cite{Li_2018_CVPR} & 74.0 & \textbf{72.4} & 4.3 & \textbf{3.9}\\
\textit{This paper} & 74.8 & \textbf{67.9} & 4.7 & \textbf{3.2}\\
\bottomrule
\end{tabular}
}
\label{tab_independent}
\vspace{-4mm}
\end{table}

%===============================
\subsection{Non-uniform kernel estimation}
%===============================
Next, we study the benefit of our proposed kernel for generating more reliable ground-truth density maps. For this experiment, we compare to the Geometry-Adaptive Kernel (GAK) of Zhang \etal \cite{zhang2016single}. For WIDER FACE, the spatial extent of objects is provided by the box annotations and we use this additional information to measure the variance quality of our kernel compared to the baseline. The counting and variance results are shown in Table \ref{tab_kernel}. The proposed kernel has a lower count error than the commonly used GAK on both ShanghaiTech Part\_A and WIDER FACE. To show that this improvement is due to the better estimation of the object size of interest, we compare the estimated variances $\sigma$ obtained by different methods with the ground truth variance obtained by leveraging the box annotations of WIDER FACE. Our kernel reduces the MAE of $\sigma$ from 2.6 to 2.2 compared to GAK.
\begin{table}[t]
\small
\caption{\textbf{Benefit of our kernel} on ShanghaiTech Part\_A and WIDER FACE. A network with our kernel obtains lower count error than with GAK \cite{zhang2016single} (see MAE~($n$) columns). To show this improvement is due to better object size estimation, we compare our kernel to the ground-truth on WIDER FACE, see MAE~($\sigma$) column, which indicates a lower size error than with GAK.}
\centering
\resizebox{0.8 \columnwidth}{!}{
\begin{tabular}{@{}lcccccccc@{}}
\toprule
 & \textbf{Part\_A}  & \multicolumn{2}{c}{\textbf{WIDER FACE}}   \\
\cmidrule(lr){2-2} \cmidrule(lr){3-4}
Kernel from & MAE ($n$) & MAE ($n$) & MAE ($\sigma$) \\
\hline
GAK~\cite{zhang2016single} & 67.9 &  4.2 & 2.6    \\
\rowcolor{Gray} \textit{This paper} & \textbf{65.2}  &  3.6 & \textbf{2.2}     \\
\midrule
Ground-truth & n.a.  &  \textbf{3.2} & n.a.     \\
\bottomrule
\end{tabular}}
\label{tab_kernel}
\vspace{-4mm}
\end{table}

%===============================
\subsection{Comparison to the state-of-the-art}
%===============================
\begin{table*}[!t]
%\small
\caption{\textbf{Comparison to the state-of-the-art for global count error} on ShanghaiTech Part\_A, Part\_B, TRANCOS, DCC, UCF-QNRF and WIDER FACE. Results on WIDER FACE based on our reimplementations. Results by Zhang \etal on UCF-QNRF taken from Idrees \etal Our results set a new state-of-the-art on all six datasets for almost all metrics.}
%\cs{Do Sam et al serve a purpose? Else remove as well.}} 
%\cs{more rows can be removed, \eg~Shi until Ranjan do not add much.}}
\centering
\resizebox{1.99 \columnwidth}{!}{
\begin{tabular}{@{}lccccccccccccr@{}}
\toprule
& \multicolumn{4}{c}{\textbf{Part\_A}} & \multicolumn{2}{c}{\textbf{Part\_B}}&\textbf{TRANCOS}&\textbf{DCC}&\multicolumn{2}{c}{\textbf{UCF-QNRF}}&\multicolumn{2}{c}{\textbf{WIDER FACE}}\\
\cmidrule(lr){2-5} \cmidrule(lr){6-7} \cmidrule(lr){8-8} \cmidrule(lr){9-9}  \cmidrule(lr){10-11} \cmidrule(lr){12-13}
 & MAE & RMSE &PSNR &SSIM & MAE & RMSE & MAE & MAE & MAE & RMSE & MAE &  NMAE\\
\hline
%Zhang \etal \cite{zhang2015cross} & 181.8 & 277.7 &-&- & 32.0 & 49.8&- &-& -&\\

%Ricardo \etal \cite{guerrero2015extremely} &-&-&-&-&-&-& 13.8 &-&-&-\\

Zhang \etal \cite{zhang2016single}& 110.2 & 173.2 & 21.4  & 0.52& 26.4 & 41.3&-& - & 277.0 & 426.0 & 7.1 & 1.10\\

%Rubio \etal \cite{onoro2016towards} &-&-&-&-&-&-& 11.0 &-&3\\

%Issam \etal \cite{Issam2018Where} & - & -&-&- & 21.6 & -&- & - &1\\

%Sam \etal \cite{sam2017switching} & 90.4 & 135.0&-&- & 21.6 & 33.4&-& - &-&-&228&445\\

%Zhang \etal \cite{Zhang_2017_CVPR} &-&-&-&-&-&-& 5.3 &-&-&\\

%Zhang \etal \cite{zhang2017fcn} &-&-&-&-&-&-& 4.2 &-&-&\\

%Liu \etal \cite{Liu_2018_CVPR_decidenet}& - & - & -&-& 20.7 & 29.4&-& - &-&-\\

Marsden \etal \cite{Marsden_2018_CVPR}& 85.7 & 131.1 &-&- & 17.7 & 28.6 &9.7 & 8.4&-&-&-&-\\

Shen \etal \cite{Shen_2018_CVPR}& 75.7 & \textbf{102.7} &-&-& 17.2 & 27.4&-&- &-&-&- &-\\

%Sindagi \& Patel \cite{sindagi2017generating}& 73.6 & 106.4& 21.7  &0.72& 20.1 & 30.1&-&- &-&-\\

%Shi \etal \cite{Shi_2018_CVPR} & 73.5 & 112.3 &-&-& 18.7 & 26.0&-& - &-&-\\

%Sam \etal \cite{Sam_2018_CVPR}& 72.5 & 118.2&-&-& 13.6 & 21.1&- & - &-&-\\

%Liu \etal \cite{Liu_2018_CVPR_Leveraging}& 72.0 & 106.6 &-&- & 13.7 & 21.4&-& - &-&-\\

%Ranjan \etal\cite{ranjan2018iterative}& 68.5 & 116.2 &-&- & 10.7 & 16.0&-& - &-&-\\

%Issam \etal \cite{Issam2018Where} &-&-&-&-&13.1&-& 3.6& -&-&-\\

Li \etal \cite{Li_2018_CVPR}& 68.2 & 115.0 &23.8  &0.76& 10.6 & 16.0&3.6 & - &- &-&4.3 &0.53 \\

Cao \etal \cite{Cao_2018_ECCV}& 67.0 & 104.5 &-&- &8.4 & 13.6&-&- &- &-&8.5 &1.10\\
Idrees \etal \cite{idrees2018composition}& - & - &-&- &- & -&-&- &132.0 &191.0 &- &-\\
\rowcolor{Gray}
\textit{This paper} & \textbf{65.2} & 109.4 & \textbf{25.4} & \textbf{0.78} & \textbf{7.2} & \textbf{12.2} &\textbf{2.0} & \textbf{3.2} & \textbf{93.8}&\textbf{146.5}& \textbf{3.2}&\textbf{0.40}\\
\bottomrule
\end{tabular}}
\label{tab_state}
\vspace{-4mm}
\end{table*}

%In the final experiment, we compare to the state-of-the-art in counting. %We first perform a comparison to standard global counting in images and then perform a comparison to counting on local level. Lastly, we evaluate the density map quality.

\textbf{Global count comparison.} 
%\psmm{I do not see a discussion on the exclusion of LIK for Shanghai Part B.} 
Table \ref{tab_state} shows the proposed approach outperforms all other models in terms of MAE on all six datasets. The proposed method achieves a new state of the art on ShanghaiTech Part\_B, and a competitive result on ShanghaiTech Part\_A in terms of RMSE. Shen \etal \cite{Shen_2018_CVPR} achieve the lowest RMSE on ShanghaiTech Part\_A, but their approach is not competitive on Part\_B. Moreover, they rely on four networks with a total of 4.8 million parameters, while our proposal just needs a single network with 2.6 million parameters. 
On TRANCOS our method reduces the count error from 3.6 (by Issam \etal \cite{Issam2018Where} and Li \etal \cite{Li_2018_CVPR}) to 2.0. A considerable reduction. For the DCC dataset proposed by Marsden \etal \cite{Marsden_2018_CVPR}, we predict a more accurate global count without any post-processing, reducing the error rate from 8.4 to 3.2. On UCF-QNRF we achieve a much better MAE and RMSE than Idrees \etal~\cite{idrees2018composition}.
For WIDER FACE, we evaluate using MAE and a normalized variant (NMAE). For NMAE, we normalize the MAE of each test image by the ground-truth face count. Again, our method achieves best results on both MAE and NMAE compared to the existing methods.

\begin{figure}[!t]
\centering
\includegraphics[width=1.05\linewidth]{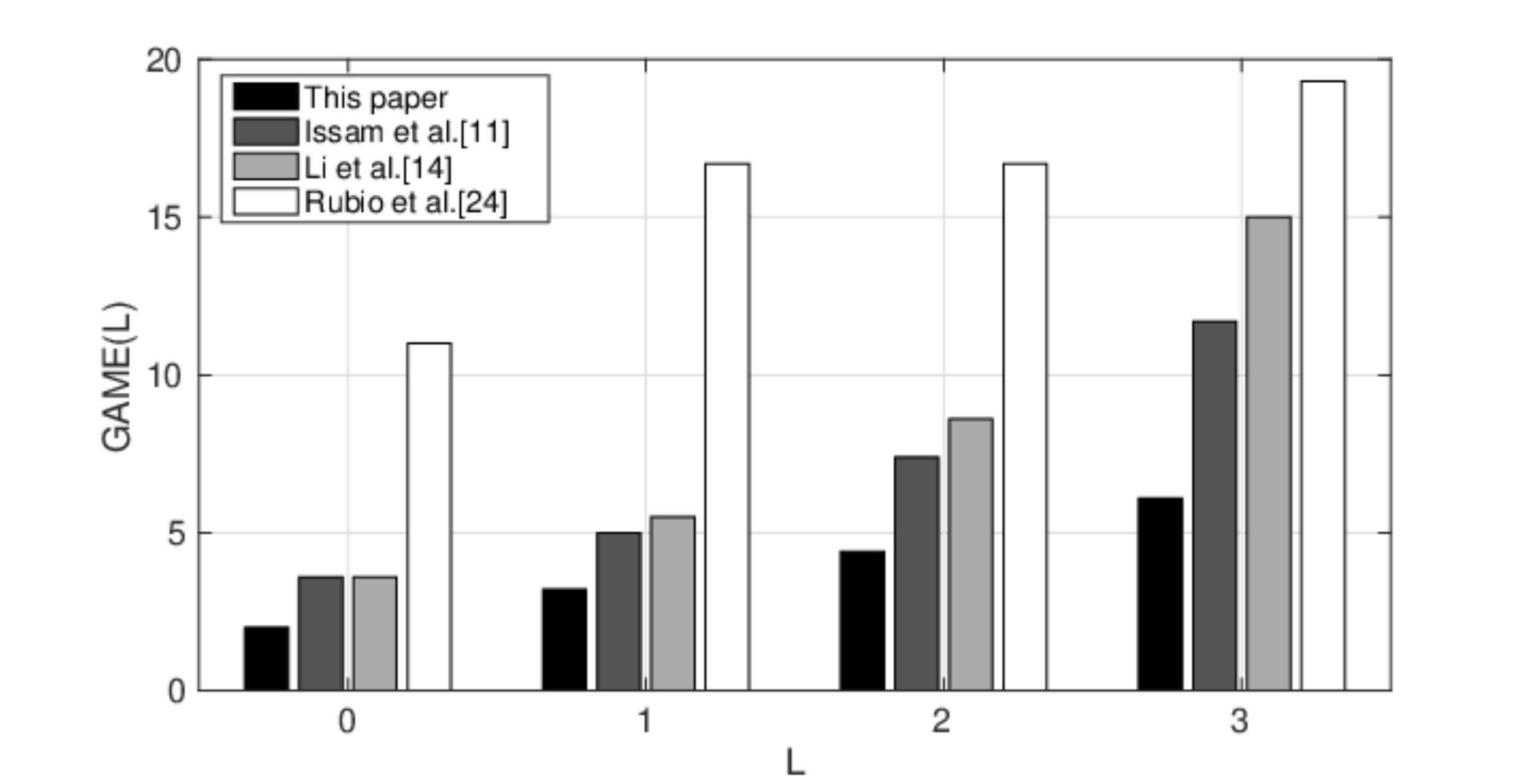}
\caption{\textbf{Comparison to the state-of-the-art for local count error} on vehicles from TRANCOS. Note the difference to other methods increases as the GAME value grows, indicating our method localizes and counts extremely overlapping vehicles more accurately.}
\label{fig_trancos}
\vspace{-6mm}
\end{figure}

\textbf{Local count comparison.} Figure \ref{fig_trancos} shows the results obtained by various methods in terms of the commonly used GAME metric on TRANCOS. The higher the GAME value, the more counting methods are penalized for local count errors. For all GAME settings, our method sets a new state-of-the-art. Furthermore, the difference to other methods increases as the GAME value increases, indicating our method localizes and counts extremely overlapping vehicles more accurately compared to alternatives.

\textbf{Density map quality.} To demonstrate that our method also generates better quality density maps, we provide results on ShanghaiTech Part\_A for the PSNR and SSIM metrics. In agreement with the results in MAE and RMSE, our method also achieves a better performance along this dimension. Compared to methods such as \cite{Li_2018_CVPR}, which produces a density map with a reduced resolution and recovers the resolution by bilinear interpolation, our method directly learns the full resolution density maps with higher quality. 

\begin{figure}[!th]
\centering
\includegraphics[width=0.99\linewidth]{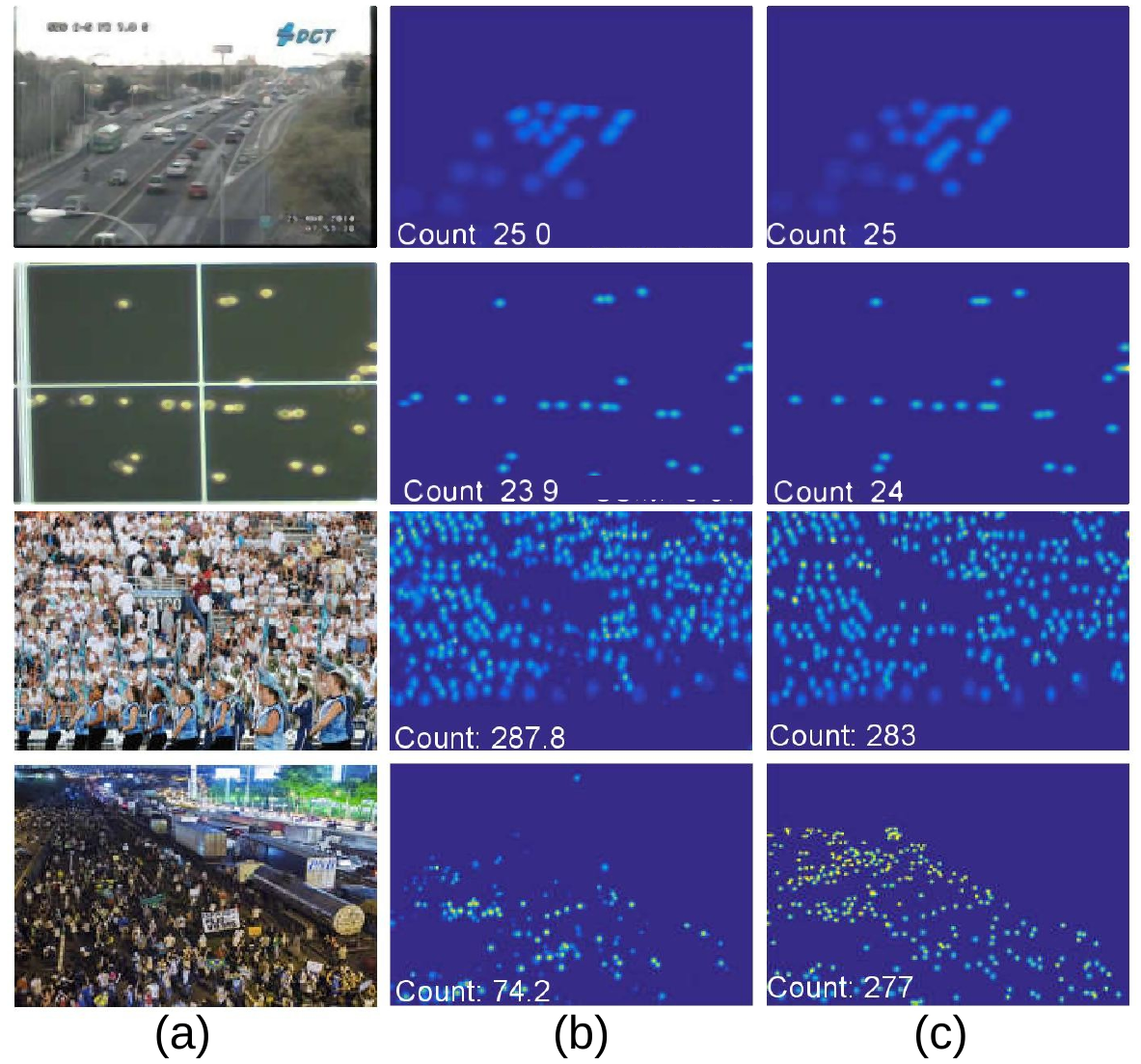}
\caption{\textbf{Density map quality.} (a) Sample images, (b) predicted density map, and (c) the ground truth. When objects are individually visible, we can count them accurately. Further improvements are required for dense settings where objects are hard to distinguish.}
\label{fig_examples}
\vspace{-6mm}
\end{figure}

\textbf{Success and failure cases.} Finally, we show some success and failure results in Figure~\ref{fig_examples}. Even in challenging scenes with relatively sparse small objects or relatively dense large objects, our method is able to achieve an accurate count (first three rows). Our approach fails when dealing with extremely dense scenes where individual objects are hard to distinguish, or where objects blend with the context (last row). Such scenarios remain open challenges.
%, we can see that scenes with extremely dense small objects or with complicated environment are still a big challenge, opening up opportunities for future work. \psmm{Explain why this goes wrong.}

%% file: conclusion.tex
\section{Conclusion}
This paper introduces two ways to repurpose the point annotations used as supervision for density-based counting. Focus from segmentation guides the counting network to focus on areas of interest, and focus from global density regularizes the counting network to learn a matching global density. Our focus for free aids density estimation from a local and global perspective, complementing each other. This paper also introduces a non-uniform kernel estimator. Experiments show the benefits of our proposal across object scales, crowding levels and base networks, resulting in state-of-the-art counting results on five benchmark datasets.
The gap towards perfect counting and our qualitative analysis shows that counting in extremely dense scenes remains an open problem. Further boosts are possible when counting is able to deal with this extreme dense scenario.

%\zl{We conclude that it is crucial for counting to regularize network to focus on object itself and avoid the disturb from context from both local and global perspective.}
%\psmm{No real new info, what insights can we give the reader in this conclusion, or next steps?}

%% file: appendix.tex
\setcounter{section}{1}
\renewcommand\thesection{\Alph{section}}
%-------------------------------
\section*{Appendix A}
%-------------------------------
In this section, we provide the architecture and ablation study of encoder-distiller-decoder network, the benefit of non-uniform kernel estimation across counting networks, and additional qualitative examples of (i) our encoder-distiller-decoder network, (ii) the effect of focus from segmentation, focus from global density and our combined focus, and (iii) success and failure cases for six benchmark datasets to better understand the benefits and limitations of the proposed method.
%-------------------------------
\subsection{Encoder-Distiller-Decoder Network}
\begin{figure*}[!t]
\centering
\includegraphics[width=0.95\linewidth]{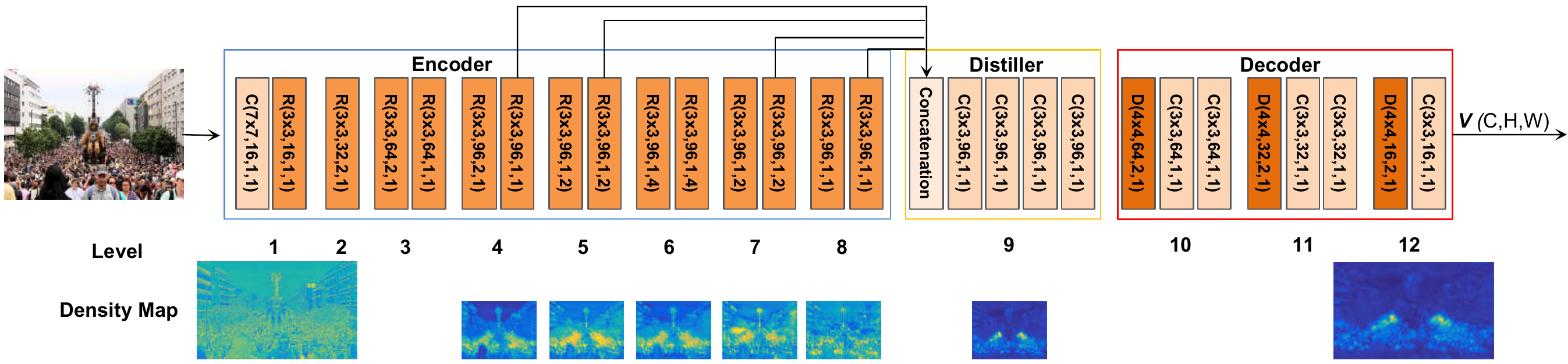}
\caption{\textbf{Encoder-distiller-decoder network}. The network consists of convolution layers (C), residual blocks (R) and deconvolution layers (D) with parameters $(k \times k, c, s, d)$, where $k \times k$ is the kernel size, $c$ is the number of channels, $s$ is the stride size and $d$ is the dilation size. Each convolution layer is followed by a ReLU activation layer and a batch normalization layer. The network is divided into several levels, such that all layers within a level have the same dilation and spatial resolution. The bottom row visualizes the mean feature map from different levels. The distiller module integrates the features from several encoder levels by attending to different parts of the image content for a better overall representation.}
\label{fig_countnet}
\end{figure*}

%-------------------------------
The proposed encoder-distiller-decoder network (Section 3.4 in the main paper) is visualized in Fig. \ref{fig_countnet}, and an ablation study on it is elaborated next.

We perform an ablation study on ShanghaiTech Part\_A to analyze the encoder-distiller-decoder network configuration. We vary the architecture by including and excluding the distiller and decoder. When relying on the encoder and distiller only, the predicted density maps are upsampled to full resolution using bilinear interpolation. Results are in Table~\ref{tab_countnet}. 

\textbf{Encoder-Distiller.} Adding a distiller module on top of the encoder reduces the MAE from 114.8 to 82.5. The distiller module fuses different features from multiple convolution layers with varying dilation rates, which is beneficial when counting multiple objects which appear in multiple scales in the image.

\textbf{Encoder-Decoder.} A traditional encoder-decoder network gives a better count than just encoder and an encoder-distiller network. An encoder-only network would compress the target objects to smaller size resulting in loss of fine details. Moreover, it produces density maps with a reduced resolution due to the downsample strides in the convolution operations. The distiller can compete with the decoder to some extent, but it cannot recover the spatial resolution and important details as well as the decoder.

\begin{figure*}[b]
\centering
\includegraphics[width=0.95\linewidth]{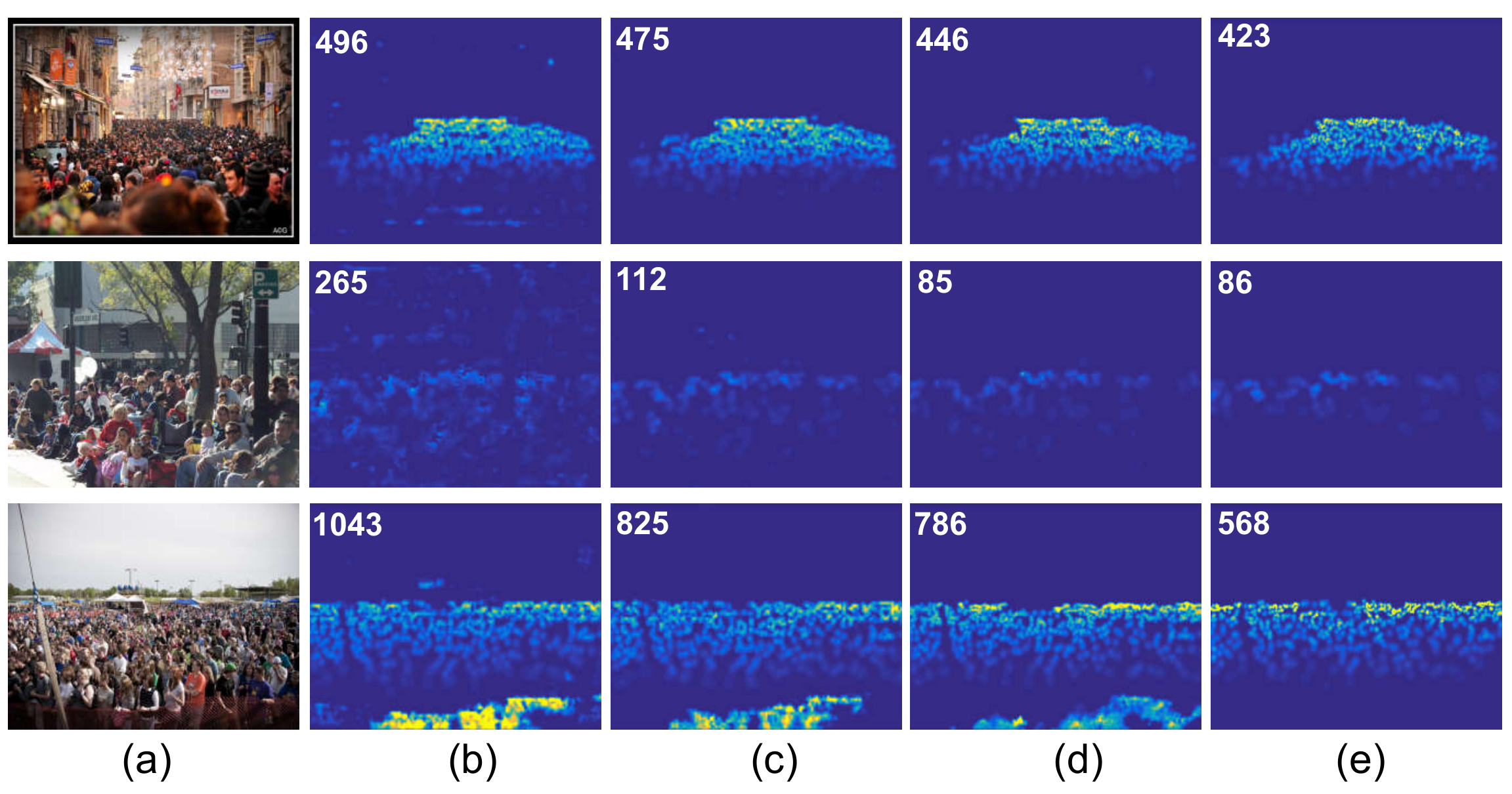}
\caption{\textbf{Ablation study of encoder-distiller-decoder network.} (a) Sample images from ShanghaiTech Part\_A and (b) predicted density map by encoder. Effect of (c) encoder-distiller and (d) encoder-distiller-decoder. For comparison, we show the ground truth for each sample in (e).}
\label{fig_ablation1}
\end{figure*}

\textbf{Encoder-Distiller-Decoder.} Incorporating the distiller in between an encoder and decoder into a single network gives the best counting results on all metrics due to the merits of both scale invariance and detail-preserving density maps. In Fig. \ref{fig_ablation1} we show qualitatively that the network obtains a lower count error and generates higher quality density maps with less noise. 

\begin{table}[!h]
\small
\centering
\caption{\textbf{Ablation study of encoder-distiller-decoder network} on ShanghaiTech Part\_A. Incorporating the proposed distiller module improves the performance of both an encoder-only network and an encoder-decoder network.}
\begin{tabular}{@{}ccccccc@{}}
\toprule
\multicolumn{3}{c}{\textbf{Encoder-distiller-decoder}} & \multicolumn{2}{c}{\textbf{Metrics}}   \\
\cmidrule(lr){1-3} \cmidrule(lr){4-5}
Encoder & Distiller & Decoder & MAE & RMSE \\
\hline
$\checkmark$&   &  &114.8 & 178.2  \\
$\checkmark$&$\checkmark$& & 82.5 & 140.6 \\
$\checkmark$&   &$\checkmark$ & 78.8 &  137.4 \\
$\checkmark$&$\checkmark$&$\checkmark$&\textbf{74.8} &\textbf{131.0}  \\
\bottomrule
\end{tabular}
%\vspace{-4mm}
\label{tab_countnet}
\end{table}

%\subsection{Qualitative Results}
%To further understand the ablation study of the encoder-distiller-decoder network, we show qualitative results in Fig. \ref{fig_ablation1}. From Fig. \ref{fig_ablation1} (b) and Fig. \ref{fig_ablation1} (c), we observe the encoder-distiller network is better able to suppress noise than the encoder-only network. This is because the distiller obtains a better overall representation for various objects by integrating the features from several encoder levels, which attend to different parts of the image content. As shown in Fig. \ref{fig_ablation1} (d), the proposed pixel-attention network (encoder-distiller-decoder) obtains a lower count error and generates higher quality density maps with less noise due to the advantages of both scale invariance and detail preservation.

%-------------------------------
\subsection{Benefit of non-uniform kernel across counting networks}
%-------------------------------
Next, we study the benefit of our non-uniform kernel estimation for existing counting methods. Apart from our own network, we also evaluate the benefit on two other counting networks, \ie~\cite{zhang2016single} and \cite{Shi_2018_CVPR}, for which code is available. Results in Table \ref{tab_kernel} demonstrate the proposed kernel has a better MAE and RMSE performance than the commonly used geometry-adaptive kernel \cite{zhang2016single} for all three networks. It demonstrates our non-uniform kernel is independent of the counting model.

\begin{table}[t]
\centering
\caption{\textbf{Benefit of non-uniform kernel estimation} on ShanghaiTech Part\_A. Relying on a ground truth density map generated by the proposed kernel, rather than GAK~\cite{zhang2016single}, lowers the counting error for our method as well as alternatives.}
\resizebox{\columnwidth}{!}{%
\begin{tabular}{lcccccc}
\toprule
 & \multicolumn{2}{c}{Zhang \etal \cite{zhang2016single}} & \multicolumn{2}{c}{Shi \etal \cite{Shi_2018_CVPR}} & \multicolumn{2}{c}{\textit{This paper}}\\
 \cmidrule(lr){2-3} \cmidrule(lr){4-5} \cmidrule(lr){6-7}
 & MAE & RMSE & MAE & RMSE & MAE & RMSE\\
\midrule
GAK~\cite{zhang2016single} & 110.2 & 173.2 & 73.5 & 112.3 & 67.9 & 115.6\\
%\rowcolor{Gray}
\textit{This paper}        & 107.0 & 156.5 & 71.7 & 109.5 & \textbf{65.2} & \textbf{109.4}\\
\bottomrule
\end{tabular}%
}
\label{tab_kernel}
\end{table}

\subsection{Qualitative Results for Segment-, Density- \& Combined-Focus}
\begin{figure*}[b]
\centering
\includegraphics[width=0.85\linewidth]{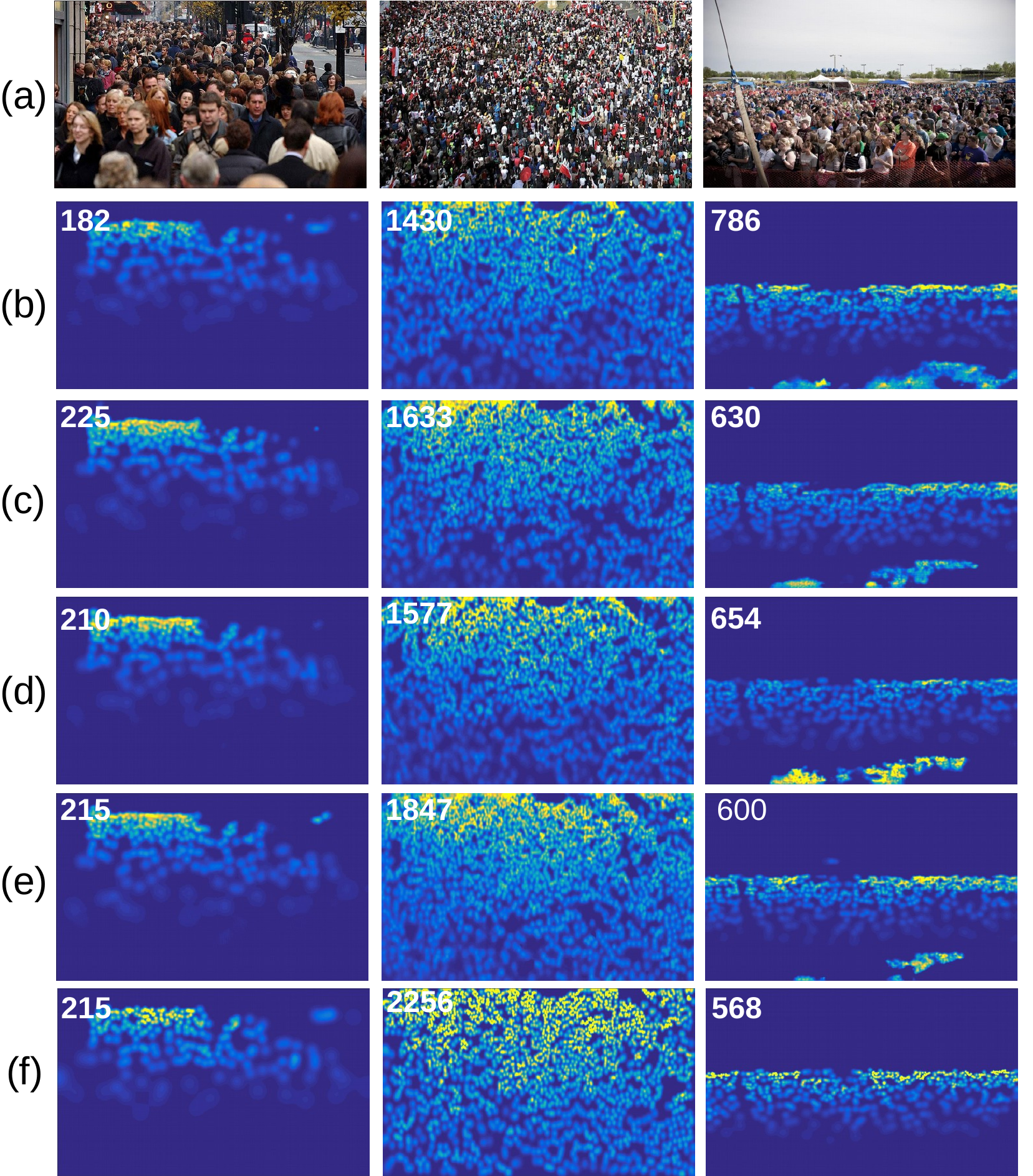}
\caption{\textbf{Effect of segment-, density- \& combined-focus} (a) Sample images from ShanghaiTech Part\_A and (b) predicted density map without focus. Effect of (c) focus from segmentation, (d) focus from global density, and (e) our combined focus. For comparison, we show the ground truth for each sample in (f).}
\label{fig_ablation2}
\end{figure*}

To illustrate the beneficial effect of the proposed focuses for reducing the counting error and suppressing background noise, we refer to Fig. \ref{fig_ablation2}. As shown in Fig. \ref{fig_ablation2} (c) and Fig. \ref{fig_ablation2} (d) compared to Fig. \ref{fig_ablation2} (b), both segmentation focus and global-density focus show the ability to suppress noise and reduce the counting error. The combination of these two focuses leads to the lowest counting error and higher quality density maps with less noise as shown in Fig. \ref{fig_ablation2} (e).

\subsection{Success and Failure Cases}

\begin{figure*}[!thbp]
\centering
\includegraphics[width=0.9\linewidth]{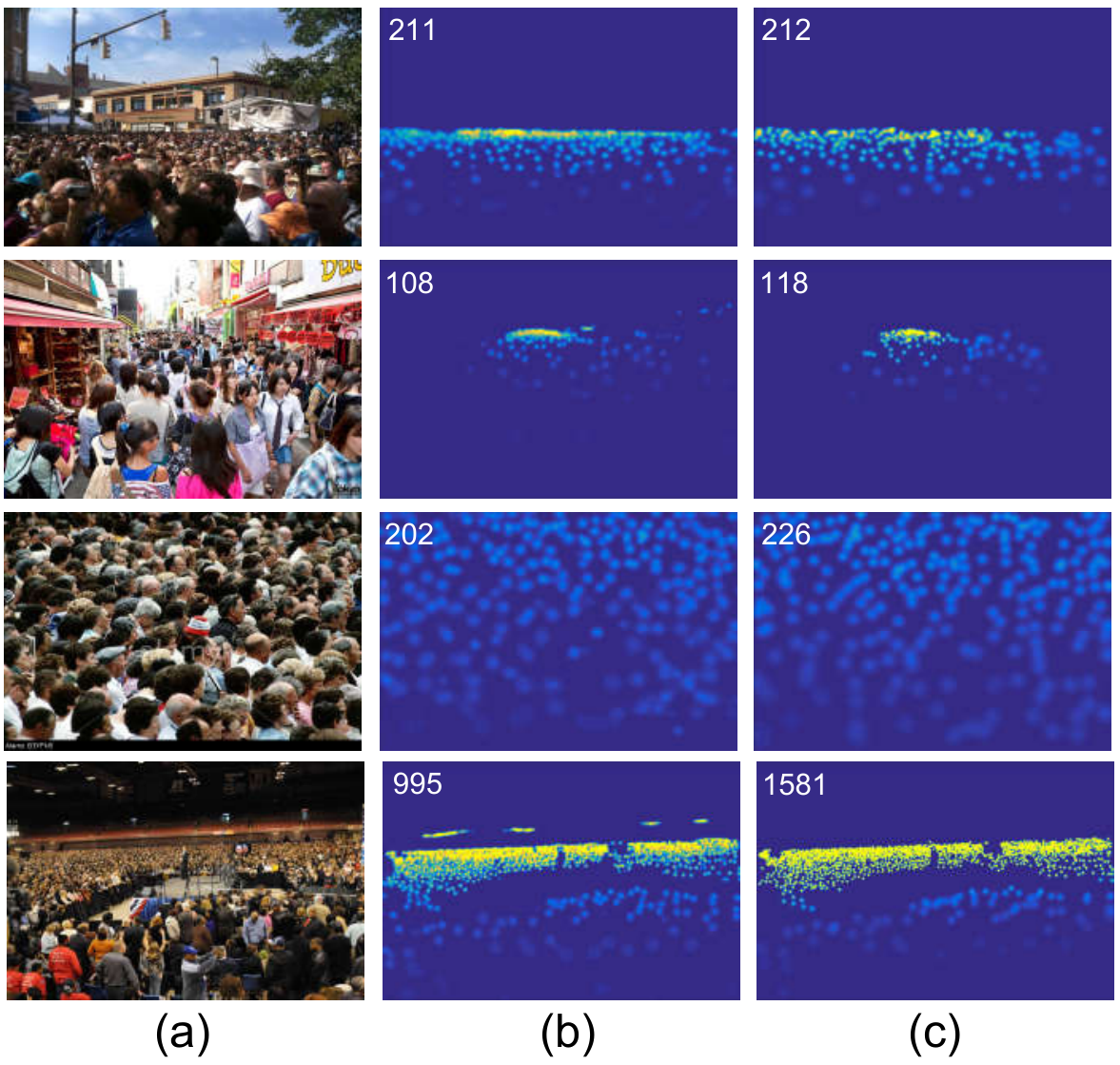}
\caption{\textbf{Qualitative results for ShanghaiTech PartA.} (a) Sample images, (b) predicted density map, and (c) the ground truth.}
\label{fig_partA}
\end{figure*}

\begin{figure*}[!thbp]
\centering
\includegraphics[width=0.9\linewidth]{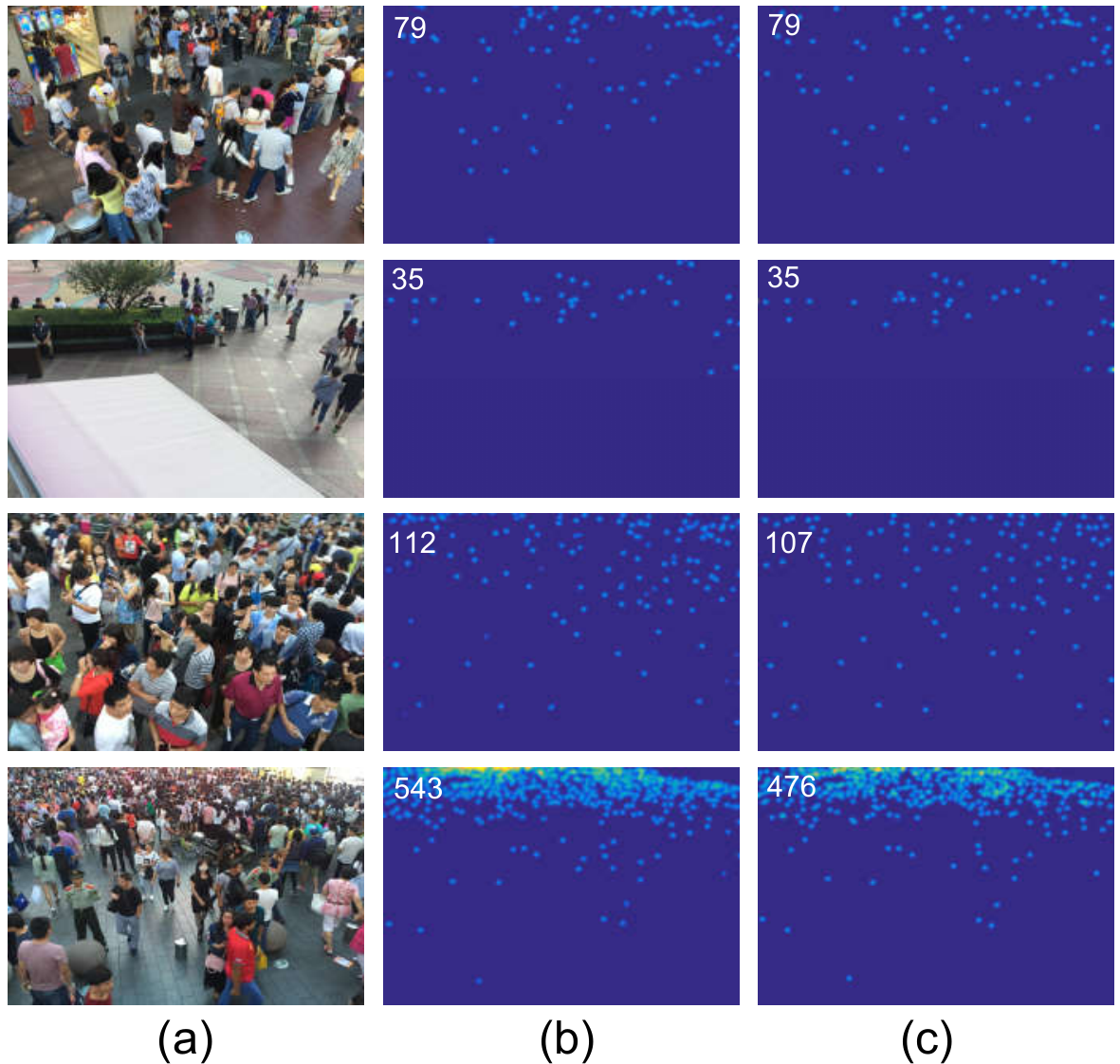}
\caption{\textbf{Qualitative results for ShanghaiTech PartB.} (a) Sample images, (b) predicted density map, and (c) the ground truth.}
\label{fig_partB}
\end{figure*}

\begin{figure*}[!thbp]
\centering
\includegraphics[width=0.9\linewidth]{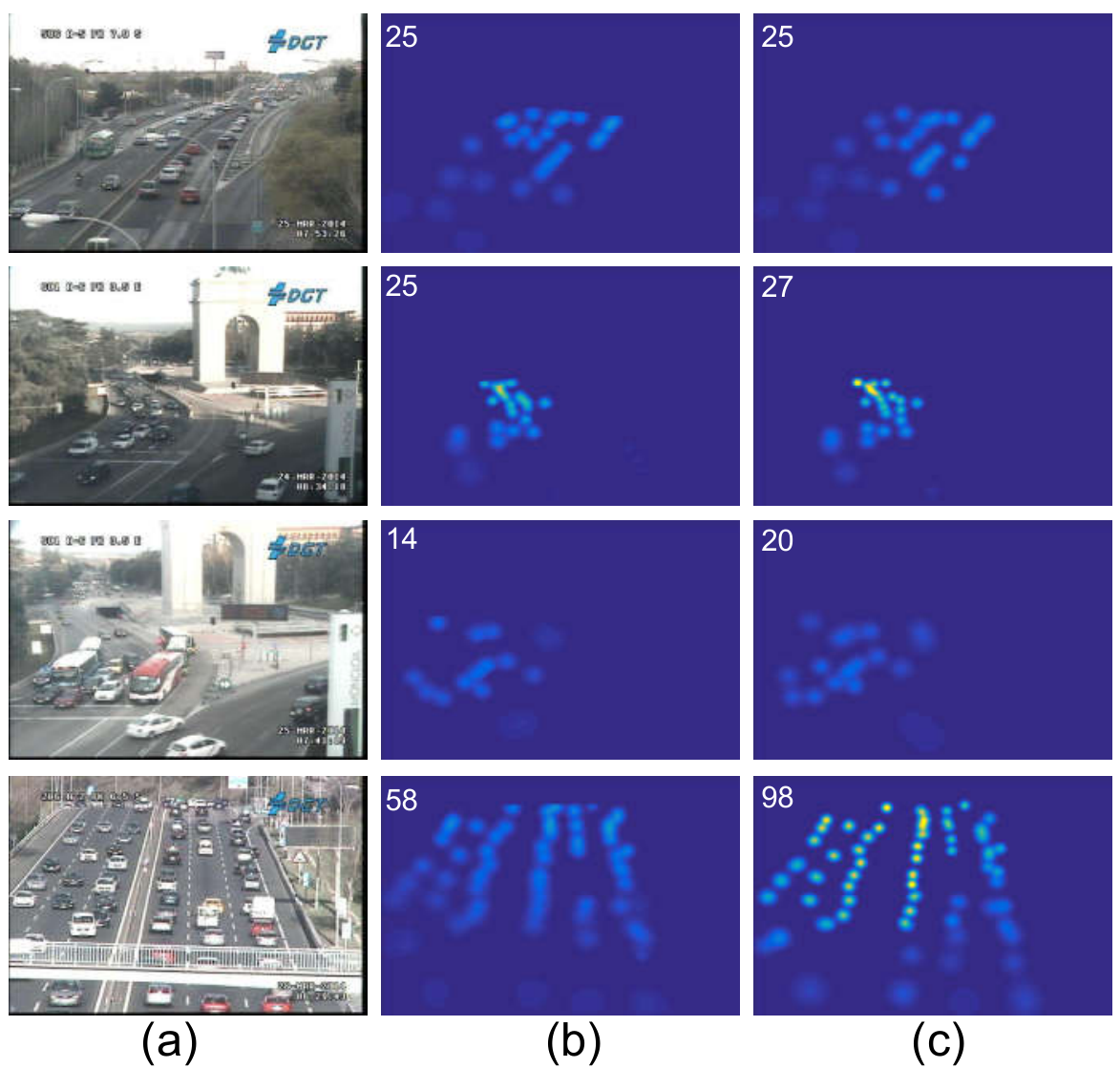}
\caption{\textbf{Qualitative results for TRANCOS.} (a) Sample images, (b) predicted density map, and (c) the ground truth.}
\label{fig_trancos}
\end{figure*}

\begin{figure*}[!thbp]
\centering
\includegraphics[width=0.9\linewidth]{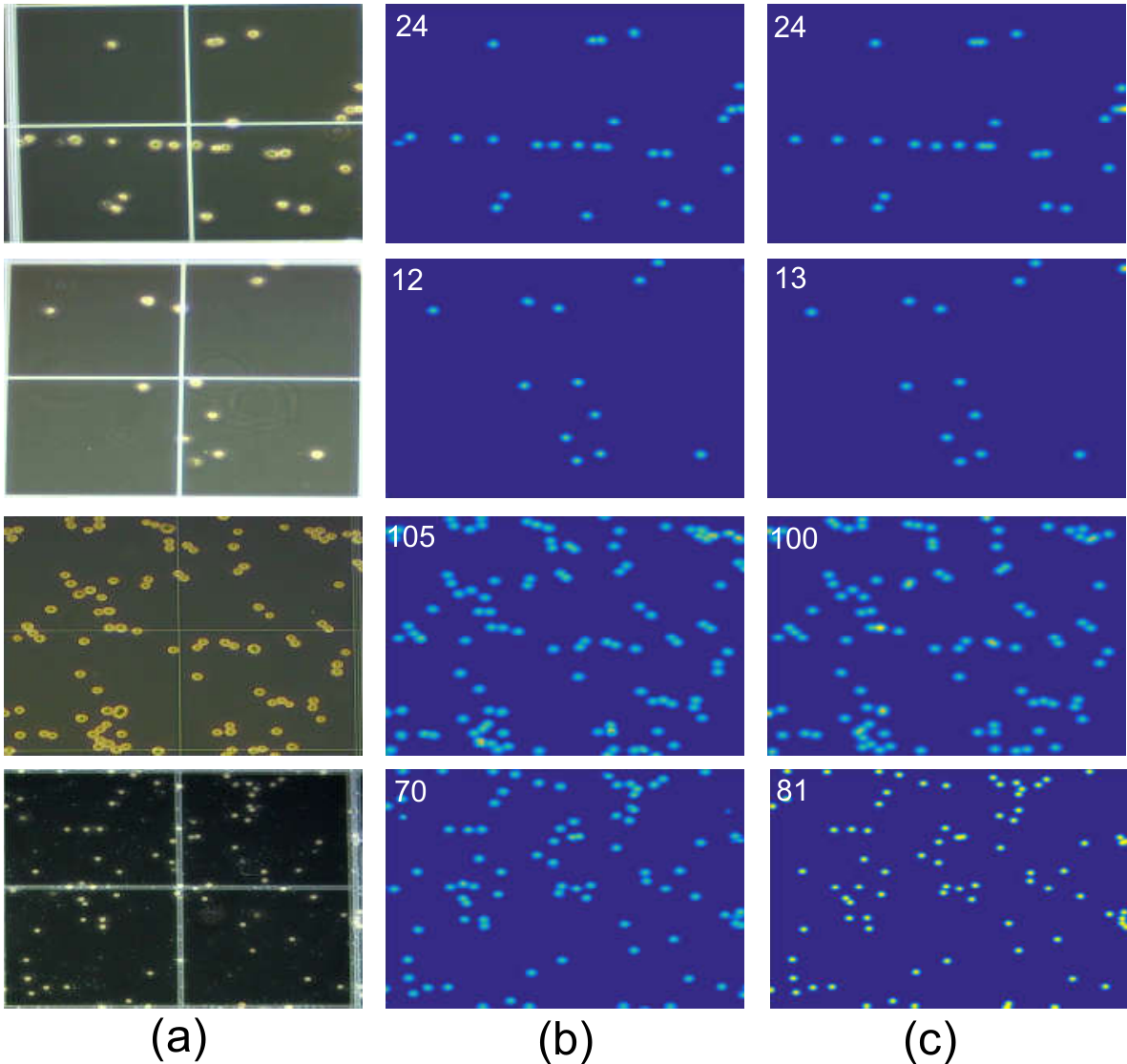}
\caption{\textbf{Qualitative results for DCC.} (a) Sample images, (b) predicted density map, and (c) the ground truth.}
\label{fig_dcc}
\end{figure*}

\begin{figure*}[!thbp]
\centering
\includegraphics[width=0.9\linewidth]{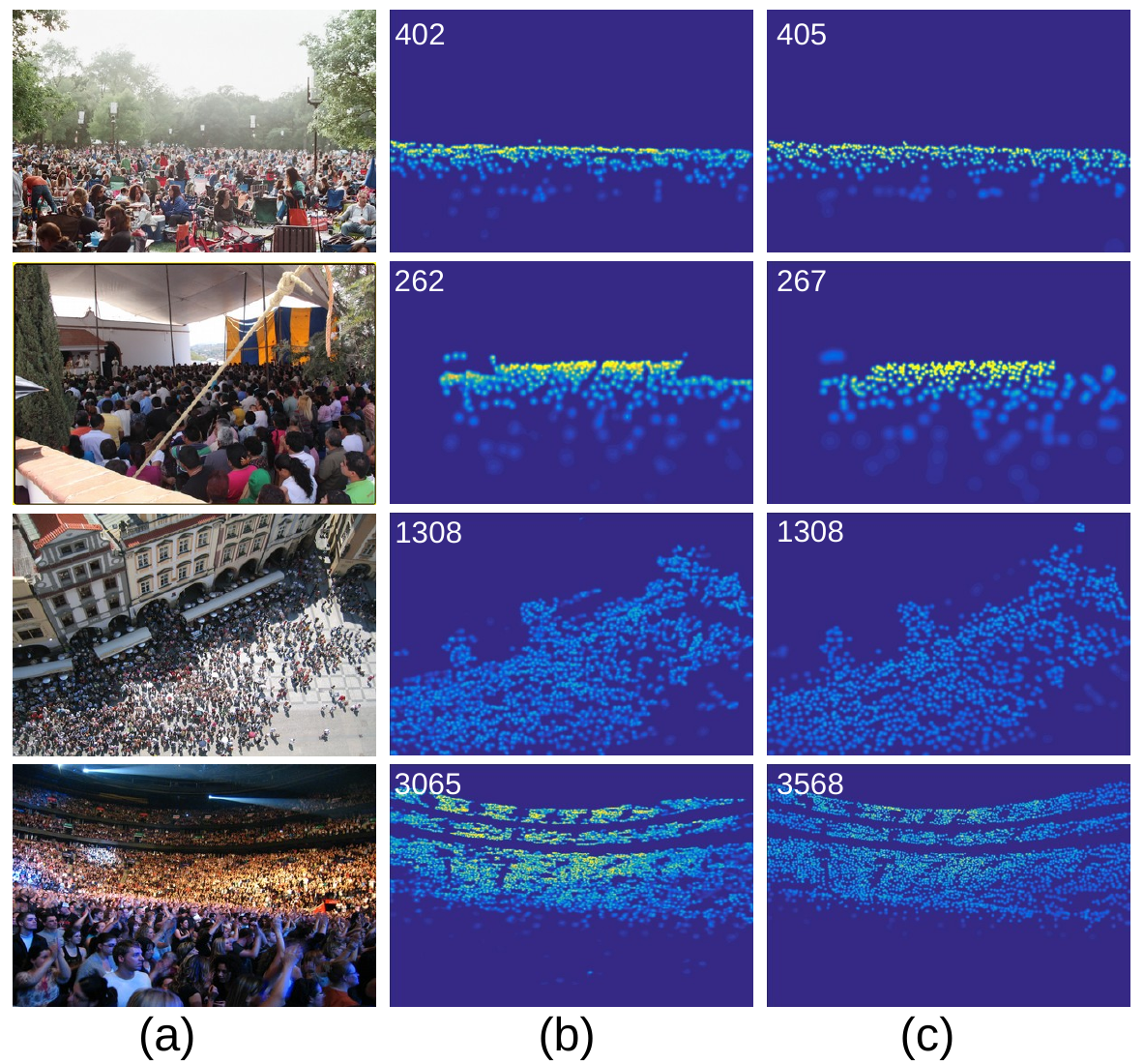}
\caption{\textbf{Qualitative results for UCF-QNRF.} (a) Sample images, (b) predicted density map, and (c) the ground truth.}
\label{fig_ucf_qnrf}
\end{figure*}

\begin{figure*}[!thbp]
\centering
\includegraphics[width=0.9\linewidth]{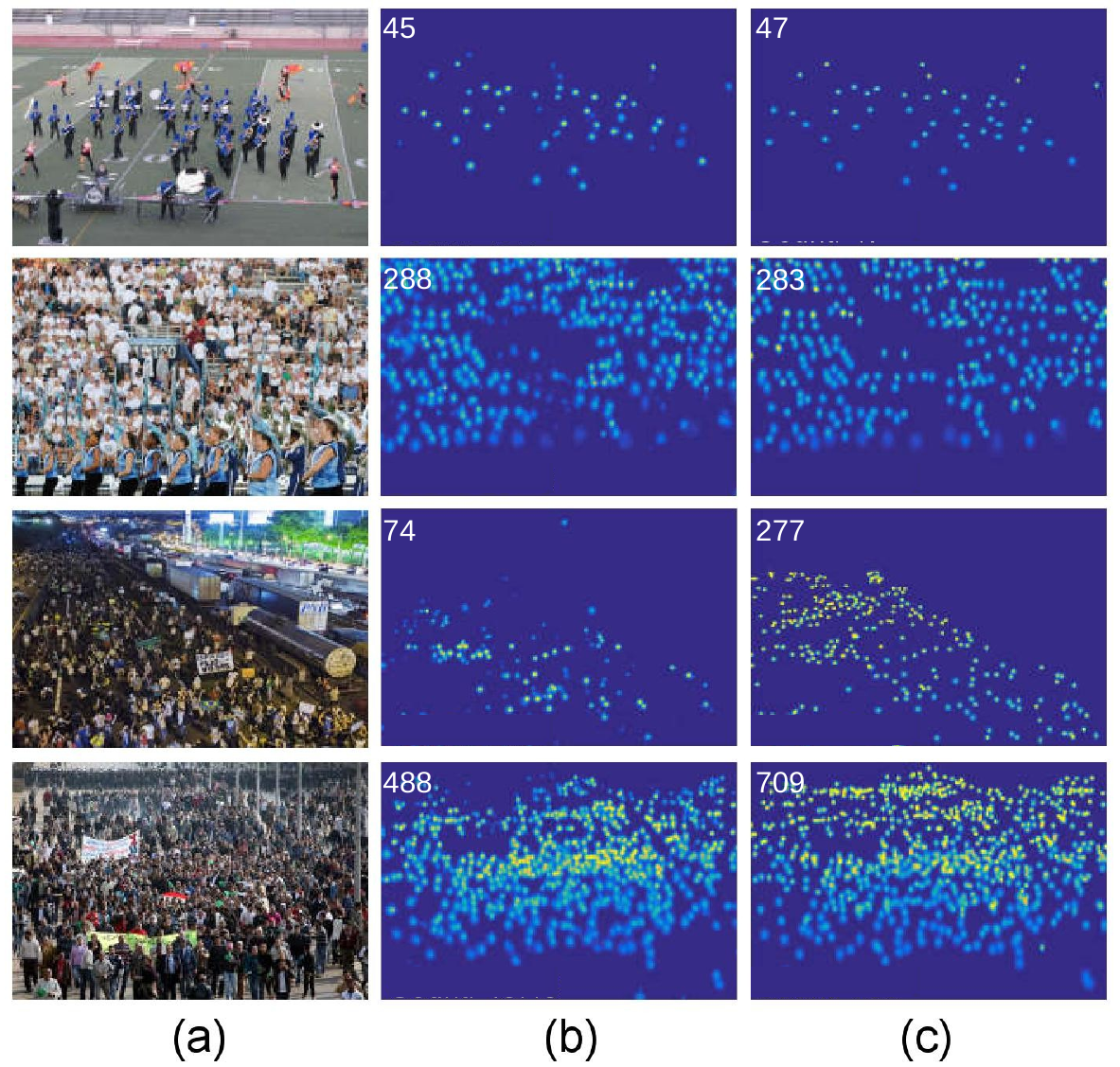}
\caption{\textbf{Qualitative results for WIDER FACE.} (a) Sample images, (b) predicted density map, and (c) the ground truth.}
\label{fig_widerface}
\end{figure*}

We have showed some success and failure results (Section 5.5 in the main paper). Finally we provide more qualitative results on all six datasets. Even in challenging scenes our method is able to achieve an accurate count, as shown in the first two rows of Fig. \ref{fig_partA}, \ref{fig_partB},  \ref{fig_trancos},  \ref{fig_dcc}, \ref{fig_ucf_qnrf} and \ref{fig_widerface}. From the failure cases, as shown in the last two rows of Fig. \ref{fig_partA}, \ref{fig_partB}, \ref{fig_trancos}, \ref{fig_dcc}, \ref{fig_ucf_qnrf} and \ref{fig_widerface}, we can see that scenes with extremely dense small objects are still a big challenge, opening up opportunities for future work.

